\documentclass[runningheads]{llncs}

\usepackage[mobile]{eccv}

\usepackage{eccvabbrv}

\usepackage{booktabs}
\usepackage{tabularx} 
\usepackage{lipsum} 

\usepackage[accsupp]{axessibility}  
\usepackage{graphicx}
\usepackage{tikz}
\usepackage{comment}
\usepackage{amsmath,amssymb} 
\usepackage{color}
\usepackage{listings}
\usepackage{booktabs}
\usepackage{multirow}
\usepackage{colortbl}
\usepackage{xcolor}
\usepackage{graphicx}
\usepackage{gensymb}
 \usepackage{adjustbox}
\usepackage{wrapfig}
\usepackage{caption}
\usepackage{comment}
\usepackage{cancel}
\usepackage{rotating} %
\usepackage{pifont}
\newcommand{\cmark}{\ding{51}}
\newcommand{\xmark}{\ding{55}}

\usepackage{multirow}

\usepackage[colorlinks,linkcolor=blue,urlcolor=blue]{hyperref}   

\usepackage{orcidlink}

\begin{document}

\title{LatentEditor: Text Driven Local Editing of 3D Scenes} 

\titlerunning{LatentEditor}

\author{Umar Khalid\inst{*,1}\orcidlink{0000-0002-3357-9720}\and Hasan Iqbal\inst{*,2}\orcidlink{0009-0005-2162-3367} \and Nazmul Karim\inst{*,1}\orcidlink{0000-0001-5522-4456} \and Muhammad  Tayyab\inst{1}\orcidlink{0000-0003-2658-2494}
  \and Jing Hua\inst{2}\orcidlink{0000-0002-3981-2933}  \and Chen Chen\inst{1}\orcidlink{0000-0003-3957-7061} 
}

\authorrunning{U.~Khalid et al.}

\institute{Center for Research in Computer Vision, University of Central Florida, Orlando, FL, USA \and
Department of Computer Science, Wayne State University, Detroit, MI, USA}

\maketitle

{ \renewcommand{\thefootnote}%
    {\fnsymbol{footnote}}
  \footnotetext[1]{* Equal Contribution}
}



\begin{abstract}
While neural fields have made significant strides in view synthesis and scene reconstruction, editing them poses a formidable challenge due to their implicit encoding of geometry and texture information from multi-view inputs. In this paper, we introduce \textsc{LatentEditor}, an innovative framework designed to empower users with the ability to perform precise and locally controlled editing of neural fields using text prompts. Leveraging denoising diffusion models, we successfully embed real-world scenes into the latent space, resulting in a faster and more adaptable NeRF backbone for editing compared to traditional methods. To enhance editing precision, we introduce a delta score to calculate the 2D mask in the latent space that serves as a guide for local modifications while preserving irrelevant regions.  Our novel pixel-level scoring approach harnesses the power of InstructPix2Pix (IP2P) to discern the disparity between IP2P conditional and unconditional noise predictions in the latent space. The edited latents conditioned on the 2D masks are then iteratively updated in the training set to achieve 3D local editing. Our approach achieves faster editing speeds and superior output quality compared to existing 3D editing models, bridging the gap between textual instructions and high-quality 3D scene editing in latent space. We show the superiority of our approach on four benchmark 3D datasets, LLFF~\cite{mildenhall2019local}, IN2N~\cite{haque2023instruct}, NeRFStudio~\cite{tancik2023nerfstudio} and NeRF-Art~\cite{wang2023nerf}. 
Project Page: \url{https://latenteditor.github.io/}
\end{abstract}
\begin{figure*}[t!]

\centering
    \includegraphics[width=0.85\linewidth, trim={3.4cm 1cm 4cm 1cm},clip]{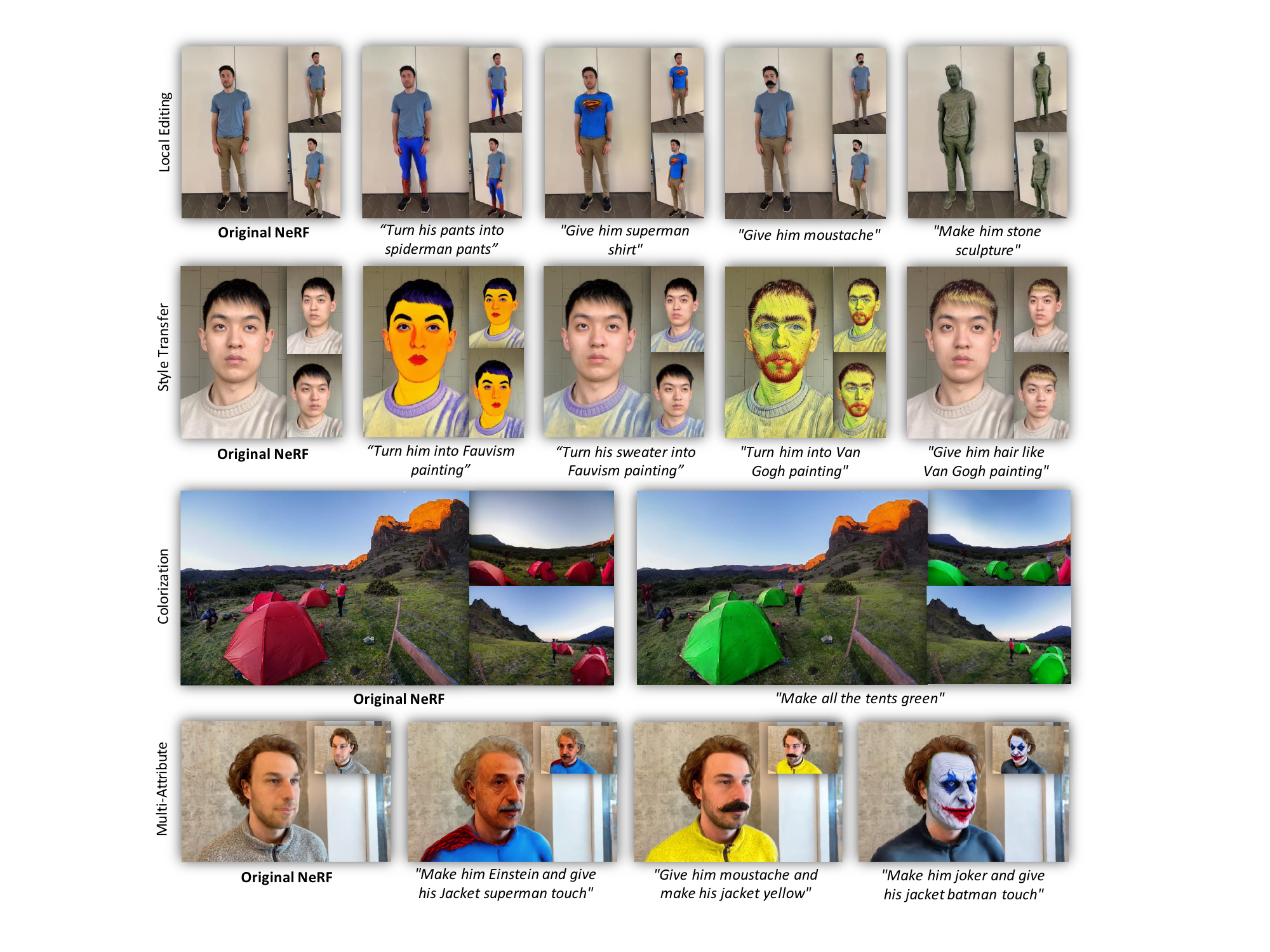}
    \caption{\footnotesize Our proposed method \textbf{\textit{LatentEditor}} enables text-based NeRF editing (e.g. color, attributes, style, etc.) by instilling both spatial and 3D awareness into image diffusion models. It can be observed in all the results that the background is intact with the color change or style transfer textual prompts. }

\label{fig:teaser}
\end{figure*}

\section{Introduction}
\label{sec:intro}
The advent of neural radiance fields (NeRF)~\cite{mildenhall2021nerf}, neural implicit functions~\cite{wang2021neus}, and subsequent innovations~\cite{liu2020neural, muller2022instant, wang2022nerf} has revolutionized 3D scene reconstruction and novel view synthesis. These neural fields, leveraging multi-view images and volume/surface rendering mechanisms, have enabled neural networks to implicitly represent both geometry and texture of scenes, offering a more reliable and user-friendly alternative to the intricate matching and complex post-processing in traditional methods~\cite{zhang2020nerf++,liu2022nerf}.

While neural fields streamline the digital representation of real-world scenes, editing these fields poses significant challenges. Unlike conventional pixel-space edits~\cite{avrahami2022blended, GLIDE, ramesh2022hierarchical, Ho20, brooks2023instructpix2pix}, editing a NeRF-represented scene requires synchronized editing of multi-views under different perspectives to maintain their consistency, a complex and error-prone task due to the high-dimensional neural network features encoding of shape and texture information. Prior approaches have explored various editing aspects~\cite{chen2021animatable, xiang2021neutex, yang2022neumesh}, yet they require extensive user interaction. More recent developments, like InstructNeRF2NeRF (IN2N)~\cite{haque2023instruct}, employ text and image-conditioned models for 3D object creation and editing. However, IN2N struggles with localizing edits.

To tackle this issue, we propose \textbf{LatentEditor} that seeks to enhance local editing by predicting the editing area from the editing prompt inside the stable diffusion~\cite{rombach2022high} denoising process. Our approach hinges on the critical insight that maintaining consistency in the diffusion feature space is viable and essential for achieving coherent edits in the 3D scene. To this end, we introduce a novel approach to assign delta scores to latent pixels in our designed delta module by contrasting noise predictions from provided instructions against unconditional noise prediction to constrain the local editing. 
However, projecting these delta scores back to the RGB pixel space introduces inconsistencies. As there is a lack of direct alignment between latent and image pixels, a slight inconsistency in the latent space can affect the NeRF training significantly. Therefore,  we propose to train NeRF directly within the latent space to enable local latent editing. 
Our approach, informed by the principles of shape-guided 3D generation in latent space~\cite{metzer2023latent}, incorporates a refining module that consists of a residual adapter with self-attention to ensure consistency between rendered latent features and the scene's original latent. During inference, LatentEditor generates a latent representation for specific poses, convertible into RGB images through a Stable Diffusion~\cite{rombach2022high} decoder. The efficacy of LatentEditor is evident in Figure~\ref{fig:teaser}, showcasing its capability for precise and minimal local edits in 3D NeRF scenes. The primary \textbf{contributions} of this paper are as follows:
\vspace{-1mm}
\begin{itemize}
  \item We introduce an efficient text-driven 3D NeRF local editing framework that operates solely based on text prompts, eliminating the need for additional controls. This innovation marks a significant advancement in text-driven 3D scene editing.
  
  \item Our unique delta module, utilizing the InstructPixtoPix~\cite{brooks2023instructpix2pix} backbone, enables a novel mechanism for local editing. Guided solely by editing prompts, it efficiently calculates 2D masks in the latent space, automatically constraining targeted modifications in precise locations.

  \item In our efficient NeRF editing method, NeRF operates directly in its latent space, reducing computational costs significantly. In addition, our designed delta module further limits the required number of editing iterations, resulting in up to a 5-fold reduction in editing time compared to the baseline IN2N~\cite{haque2023instruct}.
  
  \item A newly introduced refining module, featuring a trainable adapter with residual and self-attention mechanism, ensures enhanced consistency in the integration of latent masks within latent NeRF training. This adapter is key in aligning rendered latent features with the scene's original latents, resolving unwanted inconsistencies.
\end{itemize}

Extensive experiments on four 3D datasets and practical applications demonstrate our framework's capability to achieve spatially and semantically consistent performance and precise multi-attribute local editing in 3D scenes.

\begin{table}[t]
 \caption{\footnotesize\textbf{Comparative analysis of prior methods.} Our approach (\textit{LatentEditor}) stands out by not requiring any guidance and showcasing a broader capability for editing, particularly in achieving local edits without reliance on pre-trained segmentation models.}\label{Table1}
\scalebox{0.59}{
    \begin{tabular}{lcccccccc}
    \toprule
\multirow{2}[1]{*}{Methods}    &  \multicolumn{3}{c}{Guidance} &    \multicolumn{4}{c}{Editing Capacity}  \\
\cmidrule{2-4} \cmidrule{6-9}  &      Pre-Computed Masks & Bounding Box & GAN Guidance&    & Text Driven & Style Transfer  & Multi-Attribute Editing& Local Editing     \\
    \midrule
    {Blend-NeRF~\cite{kim20233d}}& \cellcolor{green!25}\cmark& \cellcolor{red!25}\xmark &\cellcolor{red!25}\xmark& & \cellcolor{red!25}\xmark & \cellcolor{red!25}\xmark & \cellcolor{red!25}\xmark& \cellcolor{green!25}\cmark\\
    {Blended-NeRF~\cite{gordon2023blended}}& \cellcolor{red!25}\xmark & \cellcolor{green!25}\cmark& \cellcolor{red!25}\xmark& &\cellcolor{green!25}\cmark& \cellcolor{red!25}\xmark& \cellcolor{red!25}\xmark& \cellcolor{green!25}\cmark\\
    {DreamEditor~\cite{zhuang2023dreameditor}}&\cellcolor{green!25}\cmark& \cellcolor{red!25}\xmark& \cellcolor{red!25}\xmark& &\cellcolor{green!25}\cmark&\cellcolor{green!25}\cmark& \cellcolor{red!25}\xmark&\cellcolor{green!25}\cmark\\
    {Control-4D~\cite{shao2024control4d}}& \cellcolor{red!25}\xmark& \cellcolor{red!25}\xmark&\cellcolor{green!25}\cmark&&\cellcolor{green!25}\cmark& \cellcolor{red!25}\xmark& \cellcolor{red!25}\xmark& \cellcolor{red!25}\xmark\\
    {NeRF-Art~\cite{wang2023nerf}} & \cellcolor{red!25}\xmark& \cellcolor{red!25}\xmark& \cellcolor{red!25}\xmark&   &\cellcolor{green!25}\cmark&\cellcolor{green!25}\cmark&\cellcolor{red!25}\xmark&\cellcolor{red!25}\xmark    \\
    {Instruct-N2N~\cite{haque2023instruct}} & \cellcolor{red!25}\xmark& \cellcolor{red!25}\xmark& \cellcolor{red!25}\xmark&   &\cellcolor{green!25}\cmark&\cellcolor{green!25}\cmark&\cellcolor{red!25}\xmark&\cellcolor{red!25}\xmark   \\
    \textbf{LatentEditor (Ours)} & \cellcolor{red!25}\xmark& \cellcolor{red!25}\xmark& \cellcolor{red!25}\xmark&   &\cellcolor{green!25}\cmark&\cellcolor{green!25}\cmark&\cellcolor{green!25}\cmark&\cellcolor{green!25}\cmark  \\
    \bottomrule
   \end{tabular}}
    \end{table}

\section{Related Work}
Recent advancements in denoising diffusion probabilistic models~\cite{Ho20, song2020denoising, 10.1117/12.3013575, karim2023free,iqbal2023unsupervised} have enabled high-quality image generation from complex text cues~\cite{ramesh2022hierarchical, saharia2022photorealistic, rombach2022high, karim2023free}. These models have been refined for text-driven image editing, with significant contributions by~\cite{couairon2023diffedit, kawar2023imagic, hertz2022prompt, avrahami2022blended}. Notably, SDEdit~\cite{meng2021sdedit} and DiffEdit~\cite{couairon2023diffedit} have pioneered using denoising diffusion in image editing, despite limitations in preserving original image details or handling complex captions. These advancements paved the way for novel 3D scene synthesis, particularly with text-to-image models like CLIP~\cite{clip} in DreamField~\cite{jain2022zero} and DreamFusion~\cite{poole2023dreamfusion}. 

\begin{wrapfigure}{r}{0.5\textwidth} 
    \centering
    \includegraphics[width=0.83\linewidth, trim={8cm 6cm 9cm 4cm}, clip]{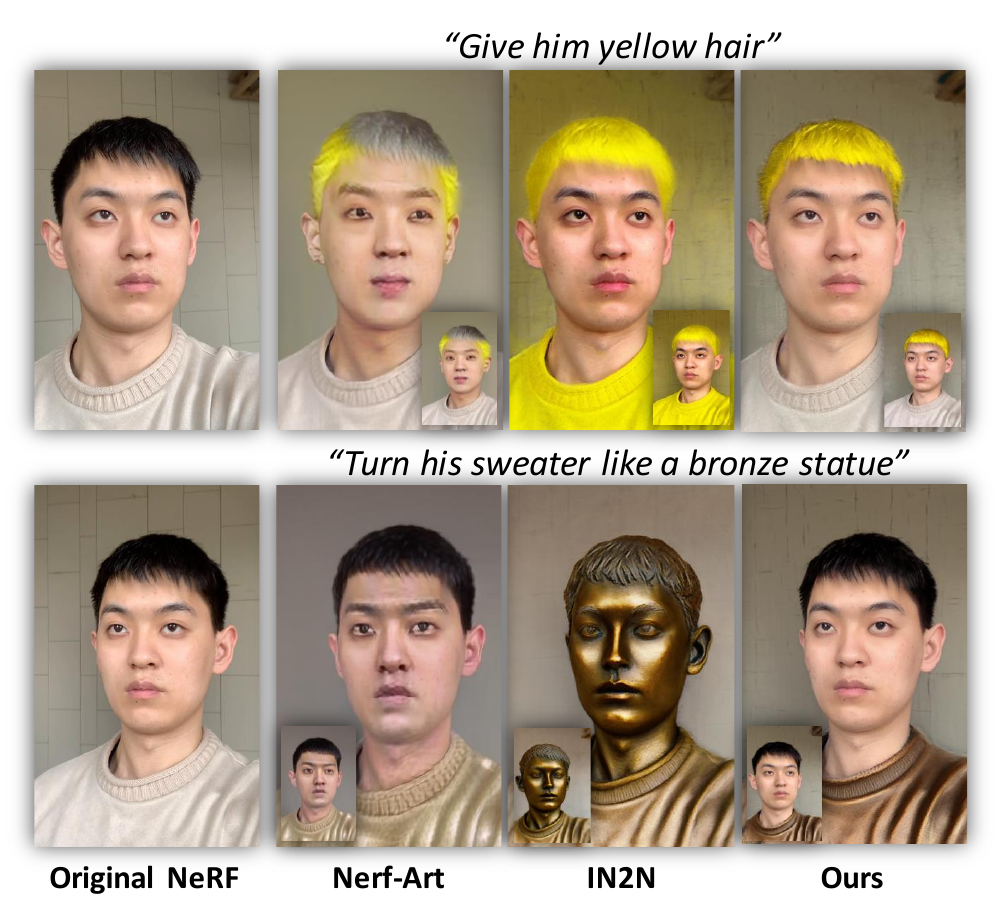} 
\caption{\footnotesize \textbf{Local Editing Challenge: }Comparative analysis of local editing capabilities between our \textit{LatentEditor} method and other text-driven 3D NeRF editing approaches, specifically IN2N~\cite{haque2023instruct} and NeRF-Art~\cite{wang2023nerf}. Our approach preserves the background seamlessly under both editing prompts.}
\label{fig:comparison}
\end{wrapfigure}
Starting with DreamFusion~\cite{poole2023dreamfusion}, subsequent methods~\cite{metzer2023latent, liu2023zero, raj2023dreambooth3d} demonstrated noteworthy outcomes using diffusion-based priors. However, these methods are limited to generating novel 3D scenes, making them unsuitable for our focus on NeRF-editing, which modifies existing 3D scenes based on provided conditions.

Compared to novel object generation, NeRF editing is less explored due to its inherent complexity. Initial efforts focused on color, geometric, and style modifications~\cite{kuang2023palettenerf, liu2021editing, wang2022clip, zhang2022arf, gu2021stylenerf}. The integration of text-to-image diffusion models is a recent trend in NeRF editing, with methodologies like Score Distillation Sampling in DreamFusion~\cite{poole2023dreamfusion}, and regularization approaches in Vox-e~\cite{sella2023vox} and NeRF-Art~\cite{wang2023nerf}. Notably, InstructNeRF2NeRF (IN2N)~\cite{haque2023instruct} utilized 2D image translation models for NeRF editing based on text prompts but faced over-editing issues. DreamEditor~\cite{zhuang2023dreameditor} addressed this with a mesh-based approach for focused edits using pre-computed masks. Similarly, Blended-NeRF~\cite{gordon2023blended} and Blend-NeRF~\cite{kim20233d} incorporated additional cues like bounding boxes for localized adjustments. Our approach innovates by integrating latent space NeRF training with a unique delta module leveraging diffusion models to generate editing masks. This enables precise local editing without extra guidance. The comparative analysis (Table~\ref{Table1} and Figure~\ref{fig:comparison}) showcases our framework's superior local editing compared to existing text-driven NeRF methods.

\begin{figure}[t]
\centering
    \includegraphics[width=0.85\linewidth, trim={0cm 0cm 0.5cm 0cm}]{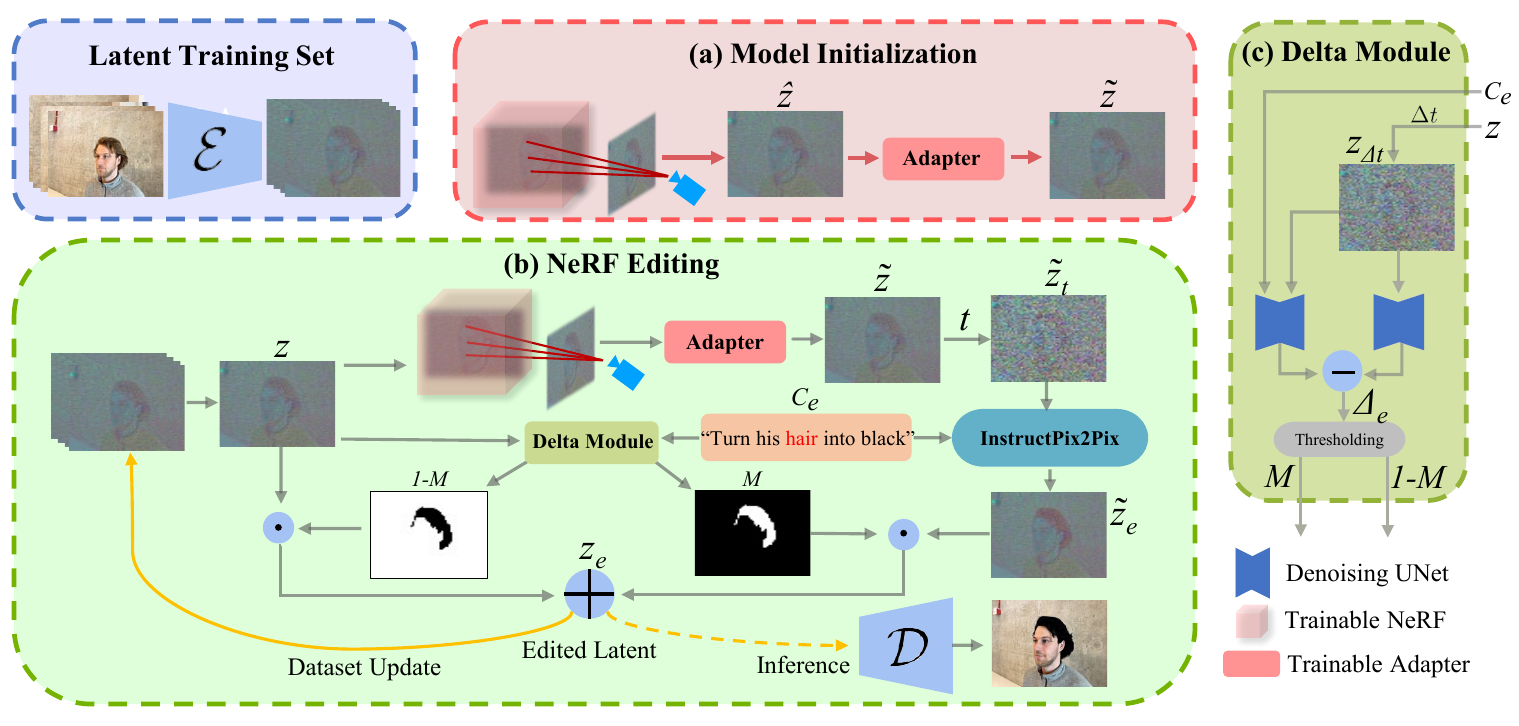}
    
    \caption{\footnotesize \textbf{Overall pipeline of LatentEditor}. (a) We initialize the NeRF model within the latent domain, guided by the latent features of the original dataset. Our refinement adapter mitigates the misalignment in the latent space and encompasses a trainable adapter with residual and self-attention mechanisms. (b) Upon initialization, LatentEditor iteratively refines the model within the latent space for a predetermined number of iterations, while consistently updating the training set with the edited latents, $Z_e$. (c) The Delta Module is adept at interpreting prompts and produces the mask for targeted editing. An RGB image can be obtained by feeding the edited latent to the stable diffusion (SD)~\cite{rombach2022high} decoder~$\mathcal{D}$ whereas $\mathcal{E}$ represents SD~\cite{rombach2022high} encoder.
 }
\label{fig:method}
\end{figure}
\section{Proposed Framework}

Our objective is to introduce a text-prompt-driven approach that enables efficient and effective local editing of real-world 3D scenes. Our method,  as illustrated in Figure~\ref{fig:method}, commences by optimizing a Neural Radiance Field (NeRF) within the latent space delineated by Stable Diffusion~\cite{rombach2022high}, thus anchoring the scene representation using latent feature vectors (see Figure~\ref{fig:method}(a)).

For scene editing, we present an innovative approach to facilitate localized editing using unconditional edits within the framework of the IP2P~\cite{brooks2023instructpix2pix} model (see Figure~\ref{fig:method}(b)). Such unconditional edits help compute latent masks within our distinctive delta module as illustrated in Figure~\ref{fig:method}(c). Original latents are edited through the guidance of the masks and prompts.

\subsection{Background}
\noindent{\textbf{InstructPix2Pix.}}
\label{sec:ip2p_prelim}
IP2P~\cite{brooks2023instructpix2pix} edits an input image $I$ based on a textual editing instruction $C_e$. Leveraging latent diffusion techniques~\cite{rombach2022high} and a Variational Autoencoder (VAE) with an encoder $\mathcal{E}$ and decoder $\mathcal{D}$, IP2P processes a noisy image latent \(z_t\). It predicts a less noisy version \(z_0\) by approximating the noise vector \(\hat{\varepsilon}\) using a U-Net \(\varepsilon_\theta\):
\begin{equation}
\hat{\varepsilon} = \varepsilon_\theta(z_t, I, C_e),
\end{equation}
where $t \in T$ represents the noise level. IP2P is trained under various conditions, enabling both conditional and unconditional denoising.

\noindent{\textbf{Neural Radiance Fields}
\label{sec:nerf_prelim}
NeRF~\cite{mildenhall2021nerf} uses Multi-Layer Perceptrons (MLPs) to estimate the density $\sigma$ and color $\mathbf{c}$ for a given 3D voxel $\mathbf{p}=(p_x, p_y, p_z)$ and view direction $\mathbf{v}$. After transforming $\mathbf{p}$ and $\mathbf{v}$ into high-frequency vectors via positional encoding $\phi(\cdot)$, NeRF produces
\begin{equation}\label{nerf_equation}
    (\mathbf{c}, \sigma) = F_{\theta}^c(\phi(\mathbf{p}), \phi(\mathbf{v})).
\end{equation}
NeRF calculates the world-space ray \(\mathbf{r}(\tau) = \mathbf{c} + \tau\mathbf{v}\) per image pixel and minimizes the difference between rendered and captured pixel colors through the loss \(\mathcal{L}(C(\mathbf{r}), \hat{C}(\mathbf{r}))\).
\vspace{1mm}

\noindent{\textbf{InstructNeRF2NeRF}
\label{sec:in2n_prelim}
IN2N~\cite{haque2023instruct} fine-tunes reconstructed NeRF models using editing instructions to create modified scenes. It employs an iterative Dataset-Update (DU) strategy, where dataset images are successively replaced with post-editing versions. This process enables the gradual integration of diffusion priors into the 3D scene. IN2N leverages an image-conditioned diffusion model, IP2P~\cite{brooks2023instructpix2pix}, to facilitate these edits.

\subsection{Method}
 We outline our latent training pipeline before delving into the details of the latent editing framework as an application stemming from latent training.

\subsection{Latent Training.}
Given a dataset of multi-view images $\mathbf{I}\in\mathbb{R}^{N\times W\times H\times 3}$, we encode these images using an encoder, $\mathcal{E}$, to obtain latent features ${z^n}=\mathcal{E}({{I^n}})\in\mathbb{R}^{W'\times H'\times 4}$ for $W'<W$ and $H'<H$. These latent feature maps $\mathbf{Z} := \{{z^n}\}_{n=1}^N$ serve as labels for initial LatentEditor NeRF training. We redefine the volume rendering integral as follows:
\begin{equation}
\hat{Z}(\mathbf{r}) = \int_{\tau_n}^{\tau_f} \mathbf{T}(\tau) \sigma(\mathbf{r}(\tau)) \mathbf{z}(\mathbf{r}(\tau, \mathbf{d})) \, d\tau,
\end{equation}
where \( \mathbf{T}(\tau) \) is the accumulated transmittance, \( \sigma(\mathbf{r}(\tau)) \) is the density, and \( \mathbf{z}(\mathbf{r}(\tau), \mathbf{d}) \) is the radiance emitted at \( \mathbf{r}(\tau) \).

The reconstruction loss, as the difference between estimated and actual pixel latent values, is defined by:
\begin{equation}\label{eq2}
\mathcal{L}_{r} = \sum_{\mathbf{r} \in \mathcal{R}} \| \hat{Z}(\mathbf{r}) - Z(\mathbf{r}) \|^2,
\end{equation}

The NeRF model $F_{\theta}^z$ is trained to predict both latent features \( \mathbf{z} \) and density \( \sigma \) from encoded positions and directions:
\begin{equation}
    (\mathbf{z}, \sigma) = F_{\theta}^z(\phi(\mathbf{p}), \phi(\mathbf{v})).
\end{equation}

\noindent{\textbf{Refinement Adapter with Self-Attention.}
Our refinement module, addressing misalignment in latent space, includes a trainable adapter for real-world 3D scene editing. It performs the following residual and self-attention operations on an input tensor $\hat{z} \in \mathbb{R}^{4 \times h' \times w'}$:
\begin{align}
    z_{\text{attention}} &= \text{SelfAttention}(\text{ConvDown}(\hat{z})) \\
    \tilde{z} &= \hat{z} + \text{ConvUp}(z_{\text{attention}}) 
\end{align}

The refinement loss for pixel latent vector $\tilde{Z}^i$ from refined feature map $\tilde{z}^i$ is:
\begin{equation}\label{eq3}
\mathcal{L}_{f} = \sum_{\mathbf{r}\in \mathcal{R}}\left \| \tilde{Z}^i(\mathbf{r}) - Z^i(\mathbf{r}) \right \|^2,
\end{equation}
where $\tilde{z}^i = F_{\Theta}(\hat{z}^i)$.

\vspace{1mm}
\noindent{\textbf{Camera Parameters Alignment in the Latent-space.}
Estimating Camera parameters from Structure-from-Motion (SfM) techniques is a standard pre-processing step for NeRF training for a scene. However, directly applying these parameters, estimated in pixel space, for NeRF training in the latent space, leads to subpar renderings marked by blurriness. While the refinement adapter mitigates this issue to a certain degree, achieving high-quality 3D reconstructions of objects and vast scenes still requires precise camera parameters for optimal outcomes. Scalar adjustments alone are insufficient for correcting camera parameter disparities which stem from the varied impact of camera parameters on image projection, influenced by the unit differences between focal length (in pixels) and translation (in world units)~\cite{park2023camp}. Facing these challenges, we propose a streamlined preconditioning strategy in the latent space for optimizing camera parameters inspired by~\cite{park2023camp}. This approach, grounded in analyzing the camera projection function's sensitivity to changes, employs a whitening transform computed via a proxy problem. This technique separates correlated parameters and balances their influence, simplifying the joint optimization process.

In our investigation of camera parameterizations, we underscore the critical impact of parameter choice on reconstruction quality. By considering camera projection as a function $\mathcal{P}^m(\phi): \mathbb{R}^k \to \mathbb{R}^{2m}$ over $m$ latnet points $\{Z_j\}_{j=1}^m$, we aim to discern the influence of camera parameters on the scene's reconstructed representation. To accurately capture the nuances of parameter effects, we examine the Jacobian matrix at a particular point $\phi^0$ in the parameter space:
\begin{align}
    \left. \frac{\partial \mathcal{P}^m}{\partial \phi} \right|_{\phi = \phi^0} = J_\mathcal{P} \in \mathbb{R}^{2m \times k},
\end{align}
where $\Sigma_\mathcal{P} = J_\mathcal{P}^\top J_\mathcal{P} \in \mathbb{R}^{k \times k}$ highlights the motion magnitude and correlation between camera parameters.
We introduce a preconditioning matrix $\mathcal{M}$ to equalize the effect of different camera parameters, ensuring that the covariance matrix $\Sigma_{\mathcal{P}'} = J_{\mathcal{P}'}^\top J_{\mathcal{P}'}$ transforms to the identity matrix $\mathcal{I}_k$. To meet this condition,  we employ Cholesky Decomposition~\cite{krishnamoorthy2013matrix}. Cholesky Decomposition provides a lower triangular matrix $\mathcal{L}$ such that:
$    \Sigma_\mathcal{P} = \mathcal{L}\mathcal{L}^\top.$
Then, the preconditioning matrix is formulated as:
\begin{align}
    \mathcal{M}^{-1} = \mathcal{L}^{-1},
\end{align}
where $\mathcal{L}^{-1}$ is the inverse of the lower triangular matrix obtained from Cholesky Decomposition of the covariance matrix $\Sigma_\mathcal{P}$. This approach aims to normalize the influence of camera parameters during optimization, improving the conditioning of the optimization problem. To implement camera parameterizations, we adopt residuals $\delta\phi_i$ applied to initial parameters $\phi_i^0$, employing FocalPose's joint pose and focal length parameterization designed for object-centric views. This setup includes principal point and radial lens distortion parameters. Originating from FocalPose's~\cite{ponimatkin2022focal} iterative pose estimation framework, we modify this parameterization to a residual form relative to initial estimates, enhancing adaptability and precision in parameter adjustments.

\noindent\textbf{Regularizing Camera Parameters.}To regularize camera parameters and prevent them from deviating significantly from their initial values during training, we introduce a regularization term to the loss function. This regularization term penalizes large deviations from the initial SfM parameter estimates, encouraging the optimization to favor solutions close to the initial values. The total loss function is defined as:
The total training loss including refinement and reconstruction losses:
\begin{equation}\label{total_loss}
\mathcal{L}_{T} = \lambda_{r} \mathcal{L}_{r} + \lambda_{f} \mathcal{L}_{f} + \lambda_{p} \mathcal{L}_{\text{reg}},
\end{equation}
We concurrently update NeRF ($F_{\theta}$) and the refinement adapter ($F_{\Theta}$) by minimizing $\mathcal{L}_{T}$ for various view reconstructions while we set $\lambda_{p}=0$ during editing where we don't optimize the camera parameters.
The regularization term can be expressed using the squared Euclidean distance (L2 norm) as follows:

\begin{equation}
\mathcal{L}_{\text{reg}} = \sum_{i} \|\theta_i - \theta_{i,0}\|^2,
\end{equation}
with $\theta_i$ denoting the current value of the camera parameter, $\theta_{i,0}$ its initial value, and the summation extending over all camera parameters. The regularization weight $\lambda$ controls the strength of the regularization, balancing the fidelity to the data against the preservation of the initial camera estimates.

\subsection{Latent Editing: An Application of Latent-NeRF}

After initializing the NeRF model in the latent domain with features \( \mathbf{Z} = \{z^n\}_{n=1}^N \), we employ the InstructPix2Pix (IP2P) framework~\cite{brooks2023instructpix2pix} to align its parameters with the textual cue \( C_{e} \). The original latent variables \( \mathbf{Z} \) are systematically replaced with edited versions \( \mathbf{Z_e} = \{z_e^n\}_{n=1}^N \) at an editing rate $\upsilon$, enabling progressive transformation to reflect desired edits. For viewpoint $K$ and editing iteration $s$, we obtain a render, $\tilde{z}^{n}$ from \( F_{\theta}^z \) and generate edited latents \( z_e^n \) using our novel local editing technique.

\noindent{\textbf{Prompt Aware Pixel Scoring.}
We design a delta module by modulating the IP2P~\cite{brooks2023instructpix2pix} diffusion process that aims to guide the generation of ${z^n_e}$ using a generated mask. Starting with noise addition to latent ${{z}^n}$ up to timestep $\Delta t$, we obtain the noisy latent ${z}^n_{\Delta t}$ as:
\begin{equation} \label{eq:delta_z}
    {z}^n_{\Delta t} = \sqrt{\beta_{\Delta t}}{z}^n + \sqrt{1 - \beta_{\Delta t}}\varepsilon,
\end{equation}
where $\varepsilon \sim \mathcal{N}(0, 1)$, and $\beta_t$ is the noise scheduling factor at timestep $t$.

IP2P's score estimation encompasses conditional and unconditional editing:

\begin{align}
    \tilde{\varepsilon}_\theta (z_t, I, C_e) &= \varepsilon_\theta(z_t, \varnothing_I, \varnothing_e) \notag \\ 
                                               &+ s_I \big(\varepsilon_\theta(z_t,  I, \varnothing_e) - \varepsilon_\theta(z_t,  \varnothing_I, \varnothing_e)\big) \notag \\
                                               &+ s_T \big(\varepsilon_\theta(z_t, I, C_e) - \varepsilon_\theta(z_t,  I, \varnothing_e)\big).
\end{align}

We calculate the delta scores, $\Delta_\varepsilon$, using two noise predictions:
\begin{equation}
\label{delta}
    \Delta _\varepsilon = \vert \varepsilon_\theta({z}^n_{\Delta t}, I, C_e) - \varepsilon_\theta({z}^n_{\Delta t},  I, \varnothing_e) \vert 
\end{equation}
where $z^n_{\Delta t}$ is calculated using Equation~\ref{eq:delta_z}, and $\Delta t $ is a hyperparameter in our method.


The higher values of the delta scores, $\Delta_\varepsilon\in\mathbb{R}^{W^{'}\times H^{'}\times 4}$ indicate the region to be edited. Hence, a binary mask, $M\in\mathbb{R}^{W^{'}\times H^{'}\times 4}$ can be generated by applying a threshold $\mu$ on $\Delta_\varepsilon$ as following,

\[
M =
\begin{cases} 
1 & \text{if } \Delta_\varepsilon \geq \mu \\
0 & \text{otherwise}
\end{cases}
\]
Given that $M$ calculated above is not unifrom across different view, we adopt a zero-shot point tracker~\cite{rajivc2023segment} to achieve consistent mask creation across all views. The process initiates by selecting query points within the first video frame's ground truth mask. These query points are determined through K-Medoids~\cite{park2009simple} sampling, which selects cluster centers as query points from the K-Medoids clustering. This strategy ensures extensive representation of different object parts and increases robustness against noise and anomalies. Since $M$ is generated in the latent space, we will perform the local editing by refining the NeRF on edited latents as discussed in the following section.

\noindent{\textbf{Local Editing with Latent Features.} Once the delta module outputs the mask $M$, a noisy version $\tilde{z}^n_{t}$ of the current render is generated, where $t$  now has a random value in a specific range $[t_{min},t_{max}]$. The edited latent  $\tilde{z}^n_{t-1,e}$  at time-step $t-1$, is computed using DDIM~\cite{song2020denoising}:
\begin{equation}
\label{denoising}
\tilde{z}_{t - 1,e}^n = \sqrt{\beta_{t - 1}} \left( \frac{\tilde{z}^n_t - \sqrt{1-\beta_t}\tilde{\varepsilon}_t}{\sqrt{\beta_t}} \right) + \sqrt{1 - \beta_{t - 1}} \tilde{\varepsilon}_t.
\end{equation}
where $\tilde{\varepsilon}_t =\tilde{\varepsilon}_\theta (\tilde{z}_t, I, C_e)$ is the predicted noise. Iteratively applying these denoising stages yields the edited latent $z_e^n$. 

The final prediction \( \tilde{z^n}_{e} \) after complete denoising merges with \({z^n} \) using the mask $M$:
\begin{equation}
\label{eq:mask.aware.prediction}
{z}^n_{e} = \tilde{z}_{e}^n  \odot M + (1 - M) \odot z^n 
\end{equation}
Ultimately, for each \( n \) in the set \( \{1, \ldots, N\} \), the locally edited latent \( z_e^n \) is substituted for the original \( z^n \) in an iterative fashion. This method ensures that pixels outside the edit mask $M$ remain unaltered, confining edits to the intended regions. 

\noindent{\textbf{Optimizing NeRF Editing.}
We don't incorporate any additional preservation or density blending losses within our methodology. The loss defined in Equation~\ref{total_loss} suffices for NeRF editing with DU, where the ground truth is continuously updated with edited latents. Our approach avoids the extensive hyperparameter tuning required by other NeRF editing methods~\cite{mikaeili2023sked,gordon2023blended,kim20233d}.

\noindent{\textbf{Multi-attribute Editing.}
We also expand our method to tackle a more challenging multi-attribute editing task in NeRF. Our approach can be seamlessly integrated with any pre-trained grounding framework, such as GLIP~\cite{li2022grounded}. However, given that the grounding mechanism is inherently embedded within our delta module, we performed prompt engineering for LLM~\cite{touvron2023llama}.

For instance, employing our designed prompt, LLM~\cite{touvron2023llama} processes the input ``Make his hair red and give him a blue jacket," yielding the output \{[``Make his hair red", ``Give him a blue jacket"], 2\}. Our delta module is designed to be aware of LLM output, generating $M^1$ and $M^2$ based on $C^1_e$ and $C^2_e$, respectively which further controls multi-edit in a single scene.      

To achieve multi-attribute editing, we leverage pre-trained large language model (LLM) Llama 2~\cite{touvron2023llama}. To achieve our multi-attribute editing, we feed in the given prompt $C_e$ to the LLM, and the LLM output is engineered in a way that our proposed delta-module can generate multiple masks.

Here is the designed prompt used in the study:
\begin{lstlisting}
### Contexts
Break the following editing prompt into multiple parts with "and" as the key indicator of partition. Produce editing prompts based on the given input.
### Input
Make his hair red and give him blue jacket.
### Response
\end{lstlisting}
\begin{figure}[t]
\centering
    \includegraphics[width=0.75\linewidth, trim={0cm 8cm 0cm 0.5cm},clip]{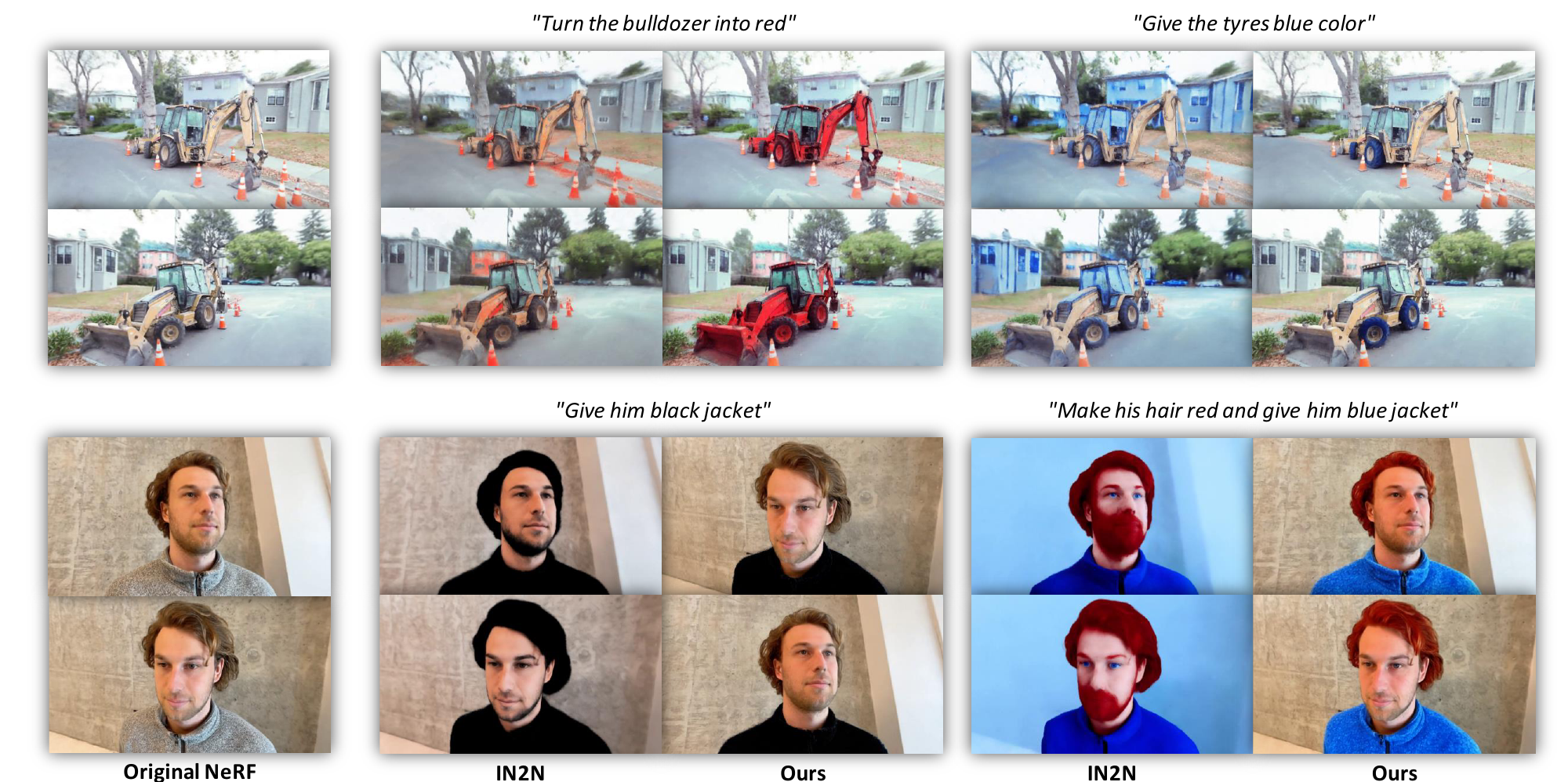}
    \caption{\footnotesize \textbf{Qualitative Results. }The visual results of our approach, when contrasted with the baseline IN2N~\cite{haque2023instruct} across two distinct scenes, distinctly demonstrate that \textit{LatentEditor} excels in accurately pinpointing the pertinent region, executing faithful text edits, and averting undesired alterations. These achievements prove challenging for the baseline method~\cite{haque2023instruct} to replicate effectively as IN2N~\cite{haque2023instruct} also changes the background objects' color to blue given that the editing prompt, "Make his hair red and give him blue jacket" only wants the jacket color to be changed.}
\label{fig:res_1}
\end{figure}
\begin{figure}[t]
\centering
    \includegraphics[width=0.75\linewidth, trim={0.7cm 8cm 0.3cm 4cm},clip]{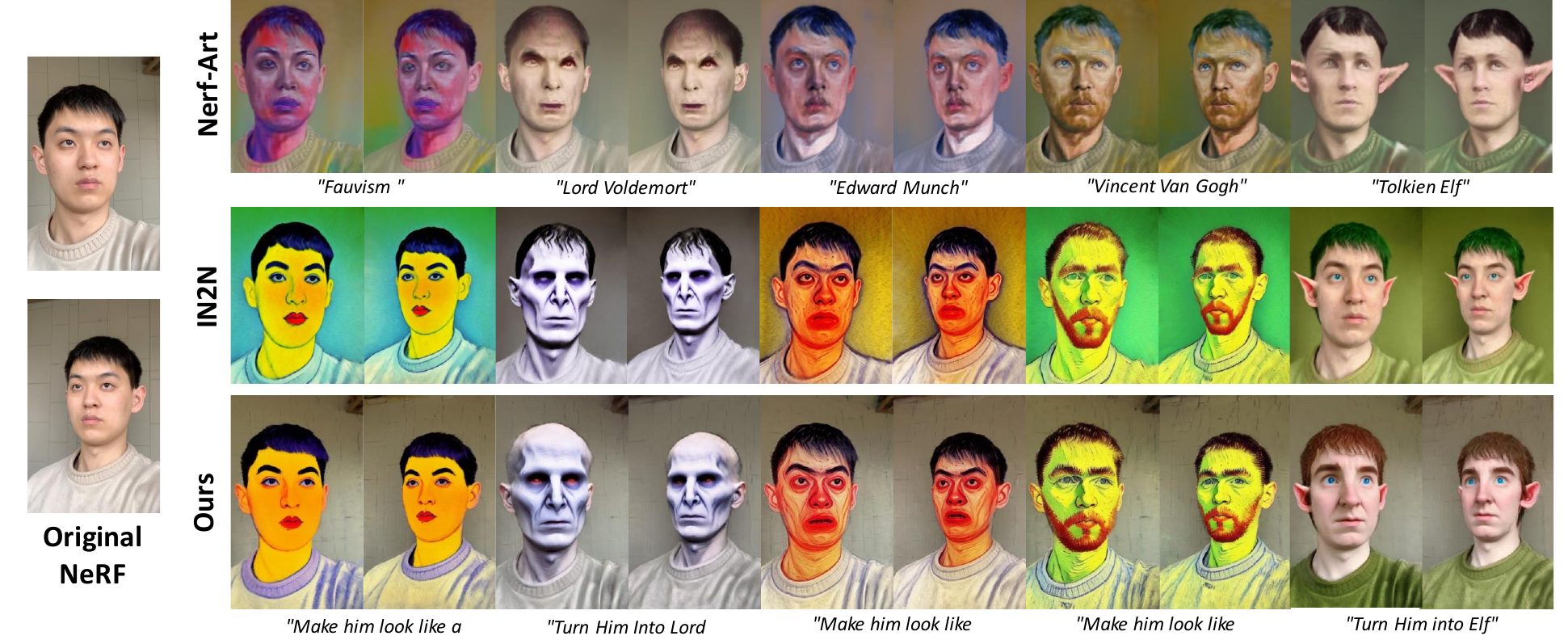}

    \caption{\footnotesize \textbf{Style Transfer Comparison. }We present a visual representation for stylization editing, comparing our results with those obtained using NeRF-Art~\cite{wang2023nerf} and IN2N~\cite{haque2023instruct}. It can be observed that LatentEditor keeps the background intact while transferring the style of an object.}
\label{fig:res_style}
\end{figure}

\section{Experimental Analysis}
\noindent\textbf{Implementation Details.} In our experiments, we adopt the implementation strategy of IN2N~\cite{haque2023instruct}, specifically setting the interval \([t_{\text{min}}, t_{\text{max}}] = [0.02, 0.98]\) and defining \(\Delta t = 0.75\). Model initialization on the original scene is performed for 30,000 iterations, ensuring a robust baseline representation. The editing process then commences, with the number of iterations tailored to the number of training views ranging from 2,000 iterations to 4,000. Detailed descriptions of these settings are available in the \textit{\textcolor{blue}{supplementary material}.}

\noindent\textbf{Baselines.}
For a comprehensive evaluation of LatentEditor's performance, we compared it against state-of-the-art (SOTA) models across four datasets: (i) IN2N \cite{haque2023instruct}, (ii) NeRF-Art \cite{wang2023nerf}, (iii) LLFF~\cite{mildenhall2019local} and (iv) NeRFstudio Dataset \cite{tancik2023nerfstudio}. We included various NeRF editing frameworks in our analysis, such as IN2N \cite{haque2023instruct}, NeRF-Art \cite{wang2023nerf}, Control-4D \cite{shao2024control4d}, and DreamEditor \cite{zhuang2023dreameditor}. However, due to space constraints, we primarily emphasize qualitative comparisons with IN2N~\cite{haque2023instruct}, the current benchmark in text-driven NeRF editing.

\subsection{Results}
\noindent\textbf{Qualitative.}
Our method's unified editing capability is distinctly showcased in Figure~\ref{fig:teaser}.Furthermore, Figure~\ref{fig:comparison} demonstrates LatentEditor's enhanced local editing prowess when juxtaposed with SOTA IN2N~\cite{haque2023instruct} approach. 
A notable distinction is observed in Figure~\ref{fig:res_1}, where our method adeptly adheres to the prompt \textit{``Give him Black Jacket"}, unlike IN2N~\cite{haque2023instruct} which erroneously alters the hair color. Although style transfer results from both LatentEditor and IN2N~\cite{haque2023instruct} are comparable, as seen in Figure~\ref{fig:res_style}, due to their shared IP2P~\cite{brooks2023instructpix2pix} backbone, LatentEditor exhibits more refined control. 

\begin{wraptable}{r}{0.5\textwidth}
\centering
\caption{\footnotesize Quantitative evaluation of scene edits in terms of text alignment and frame consistency in CLIP space.}
\resizebox{\linewidth}{!}{%
\begin{tabular}{lccc}
\hline
\textbf{Method} &
  \textbf{\begin{tabular}[c]{@{}c@{}}CLIP Text-Image\\ Direction Similarity$\uparrow$\end{tabular}} &
  \textbf{\begin{tabular}[c]{@{}c@{}}CLIP Direction\\ Consistency$\uparrow$\end{tabular}} &
  \textbf{\begin{tabular}[c]{@{}c@{}}Edit\\ PSNR$\uparrow$\end{tabular}} \\ \hline
\textbf{NeRF-Art}~\cite{wang2023nerf} & 0.2617 & 0.9188 & 21.04 \\
\textbf{Control4D~\cite{shao2024control4d}} & 0.2378 & 0.9263 & 19.85 \\
\textbf{DreamEditor~\cite{zhuang2023dreameditor}} & 0.2474 & 0.9312 & 20.67 \\
\textbf{IN2N}~\cite{haque2023instruct} & 0.2649 & 0.9358 & 24.07 \\
\textbf{Ours}             & \textbf{0.2661} & \textbf{0.9387} & \textbf{25.15} \\ \hline
\end{tabular}%
}
\label{tab:tab2}
\end{wraptable}

The versatility of LatentEditor in managing complex prompts and multi-attribute edits is further highlighted in Figure~\ref{fig:res_1}, underscoring its robustness in diverse editing scenarios. We also assess our technique in \textbf{object removal}, aiming to seamlessly erase objects from 3D scenes, which might result in voids due to missing data. This process involves detecting and excising areas using the 2D mask, and then repairing these "invisible regions" in the original 2D frames with LAMA ~\cite{suvorov2022resolution}. As shown in Figure~\ref{fig:removal}, LatentEditor excels in removing objects, delivering superior spatial accuracy and view consistency compared to Gaussian Grouping~\cite{ye2023gaussian}.






\begin{figure}[t!]
\centering
    \includegraphics[width=0.75\linewidth,trim={0cm 0cm 0cm 0cm}, clip]{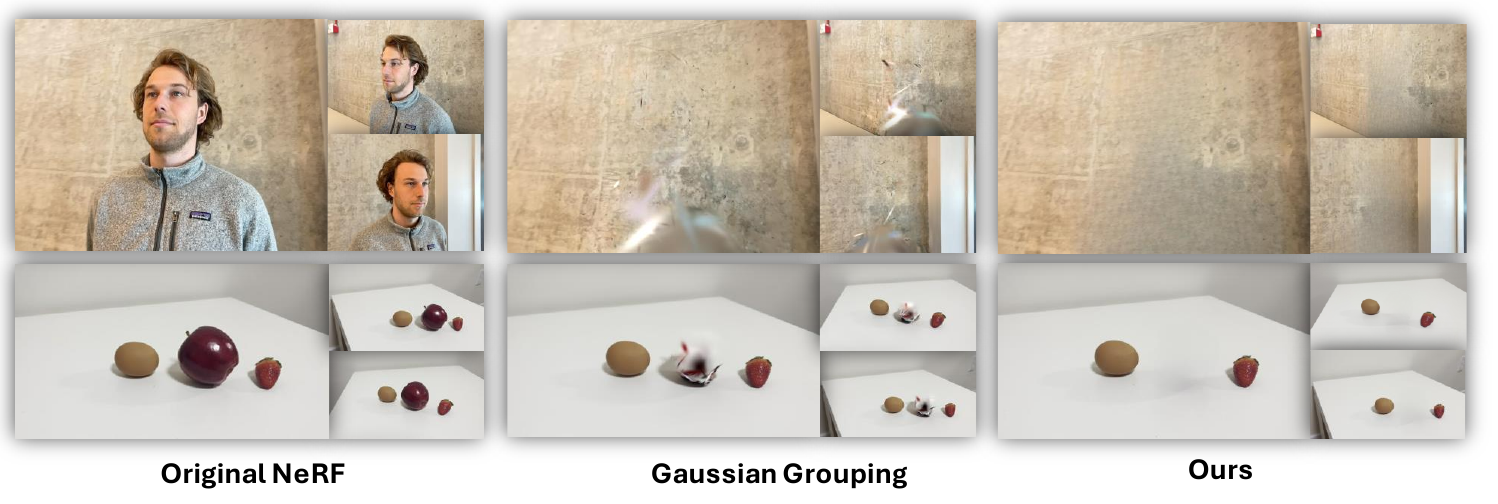}

\caption{\footnotesize\textbf{Object Removal. }Our method outperforms Gaussian Grouping~\cite{ye2023gaussian} in removing 3D objects across various scenes.}
\label{fig:removal}
\end{figure}

\noindent\textbf{Quantitative.}
Quantitative evaluations, as detailed in Table~\ref{tab:tab2}, 200 edits across 20 scenes from the above-mentioned four datasets. Our method outperformed baselines in CLIP similarity scores and CLIP direction consistency, as averaged over multiple views rendered from NeRF.  

\begin{wrapfigure}{r}{0.5\textwidth} 

    \centering

    \includegraphics[width=0.98\linewidth,trim={0.2cm 0cm 0.2cm 0.6cm}, clip]{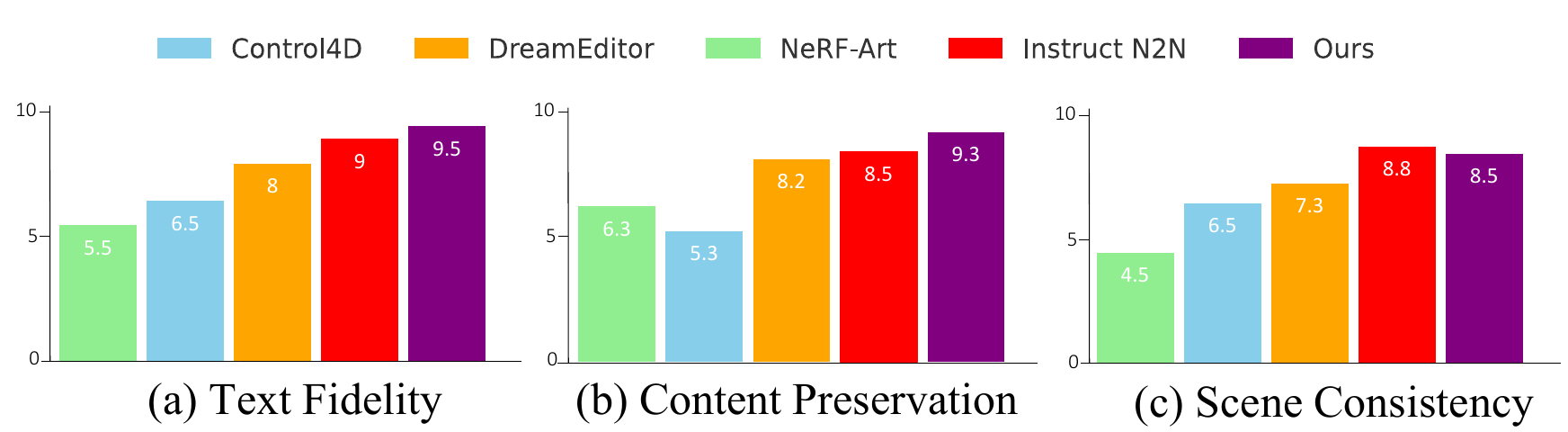}

\caption{\footnotesize\textbf{User Study. } LatentEditor achieves the highest text fidelity and content preservation scores.}

\label{fig:user_study}

\end{wrapfigure}
\noindent\textbf{User Study.}
To evaluate the subjective nature of scene editing, we conducted a user study comparing our method with SOTA alternatives. The study garnered a total of 500 votes across three key metrics: 3D consistency, preservation of original scene content, and adherence to text descriptions. As depicted in Figure~\ref{fig:user_study}, our method has been predominantly favored across these metrics. Further details on the quantitative evaluation criteria and implementation of this user study are available in the \textit{\textcolor{blue}{supplementary material}.}

\section{Ablations}
\noindent{\textbf{Editing Rate vs Training Iterations.}}
In this ablation study, we evaluate the computational efficiency of our LatentEditor approach against IN2N~\cite{haque2023instruct}, particularly in terms of training iterations required to achieve a targeted editing performance. In Figure~\ref{fig:ablation_1}, using the editing prompt \textit{``Turn his hair black"}, we demonstrate that LatentEditor significantly outperforms IN2N~\cite{haque2023instruct} in computational cost.

Specifically, our method achieves the desired editing results with an approximately five to ten-fold reduction in training iterations. LatentEditor requires only about 2000 iterations to reach the editing rate benchmark, while IN2N~\cite{haque2023instruct} still faces challenges even after 10,000 iterations. Despite LatentEditor involves multiple denoising stages per editing step, which increases the cost per step, it achieves the desired performance with significantly fewer iterations. As a result, it is \textbf{5-8} times faster in NeRF editing compared to IN2N~\cite{haque2023instruct}.

\begin{figure}[t]
\centering
    \includegraphics[width=0.75\linewidth, trim={0cm 0.8cm 0cm 0cm},clip]{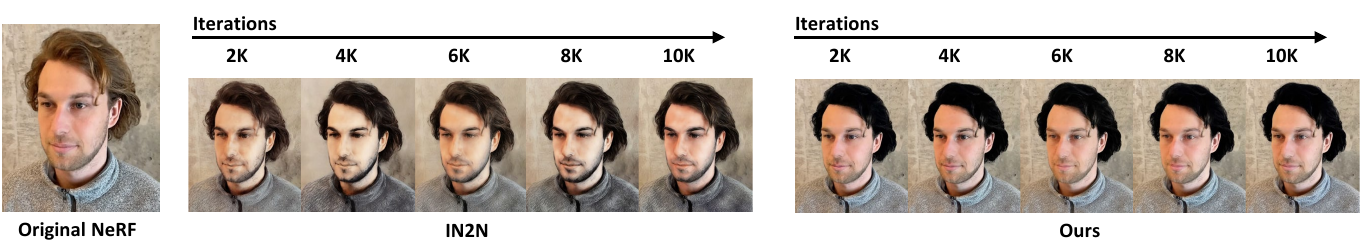}
    \caption{\footnotesize Comparing \textit{LatentEditor} to IN2N in terms of a reduced computational cost. Our approach achieves the desired editing results, ``Turn his hair black", in approximately 2000 iterations, whereas IN2N continues to face challenges even after 10,000 iterations.}
\label{fig:ablation_1}
\end{figure}

\begin{wrapfigure}{r}{0.5\textwidth} 
    \centering
  \
    \includegraphics[width=0.48\textwidth, clip]{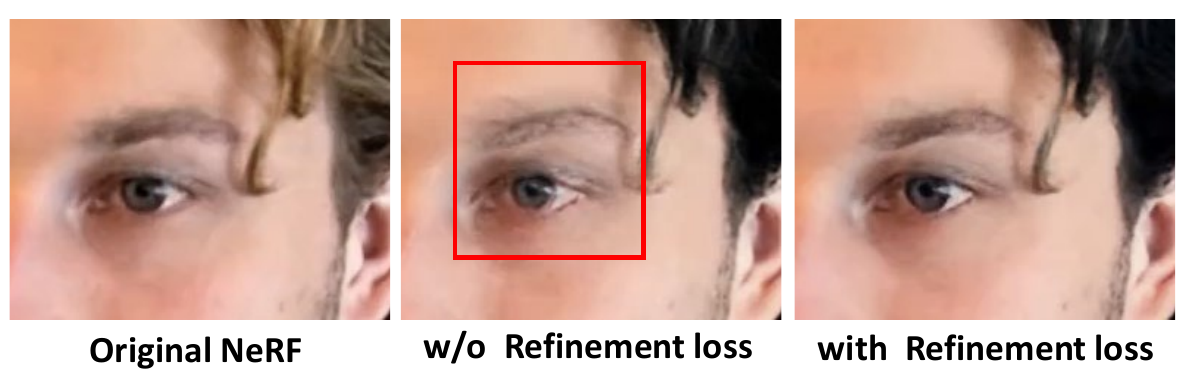} 
\caption{\footnotesize Qualitative comparison of an edited image with and without refinement loss against the editing prompt, \textit{``turn his hair black"}. The red box indicates the noticeable artifacts without the refinement module.}
\label{fig:ablations_2}

\end{wrapfigure}
\noindent{\textbf{Refinement Coefficient Sensitivity.}
 To understand the impact of the refinement module, we conduct an ablation study on the refinement coefficient where this coefficient is set to zero. The results, as illustrated in Figure~\ref{fig:ablations_2}, reveal that omitting the refinement leads to noticeable inconsistencies and artifacts in the NeRF scene. These findings highlight the importance of the refinement loss in achieving high-quality, consistent NeRF scenes and validate its inclusion in our loss formulation.

\section{Limitations}
Our method's efficacy is contingent on the capabilities of the pre-trained IP2P model~\cite{brooks2023instructpix2pix}, which presents certain limitations. This is particularly evident in cases where IP2P's inherent weaknesses are pronounced. For instance, in Figure~\ref{fig:limitations}, the prompt \textit{``Turn the bear into an orange bear"} exemplifies such a scenario. While IN2N~\cite{haque2023instruct} introduces random coloring throughout the scene, failing to generate the desired NeRF, our method, though demonstrating more controlled editing, does not completely succeed in turning the bear orange. The underlying limitation stems from IP2P's challenges in accurately interpreting and executing specific editing instructions like this. Our approach, being model-agnostic, can benefit from future enhancements in instruction-conditioned diffusion models, potentially overcoming these current constraints in localized edits.
 \begin{wrapfigure}{r}{0.5\textwidth} 
 
    \centering
    \includegraphics[width=0.48\textwidth,  trim={1cm 7.5cm 1cm 7cm},clip]{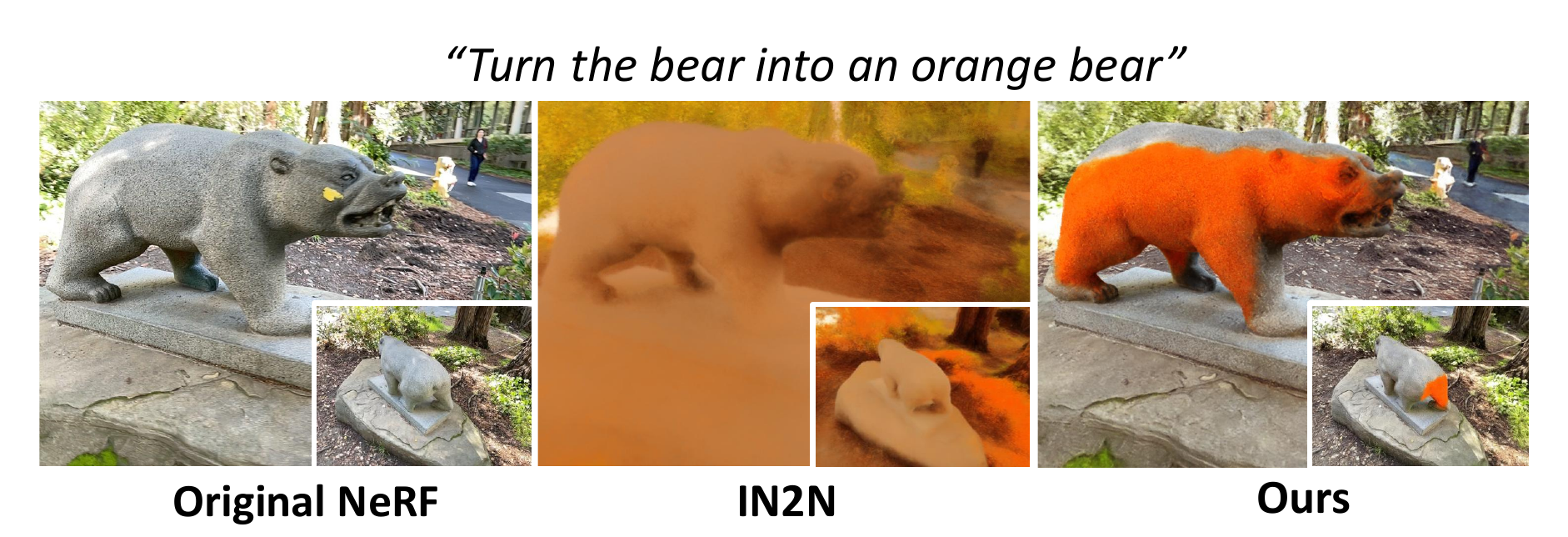} 
\caption{\footnotesize Our method grapples with instructions like ``Turn the bear into an orange bear'' due to IP2P's limitations, while LatentEditor's model-agnostic approach offers promise for addressing such issues through enhanced instruction-conditioned diffusion models.}
\label{fig:limitations}
 
\end{wrapfigure}

\section{Conclusion}

In conclusion,  LatentEditor marks a significant advancement in the field of neural field editing. We tackled the inherent challenges in editing neural fields, which arise from their implicit encoding of geometry and texture, by introducing a novel framework capable of precise, controlled editing via text prompts. By embedding real-world scenes into latent space using denoising diffusion models, our framework offers a faster and more adaptable NeRF backbone for text-driven editing. The introduction of the delta score, which calculates 2D masks in latent space for precise editing while preserving untargeted areas, is a key innovation. 
 LatentEditor not only simplifies the process of 3D scene editing with textual instructions but also enhances the quality of the results, marking a new direction in 3D content creation and modification.

\section{Acknowledgement}
This work was partially supported by the NSF under Grant Numbers OAC-1910469 and OAC-2311245.


\clearpage  

%
%
\bibliographystyle{splncs04}
\bibliography{egbib}
\end{document}


\maketitle
 \setcounter{section}{0}
  \setcounter{table}{0}
   \setcounter{figure}{0}
\renewcommand{\thesection}{\Alph{section}}

\section{Overview}
The supplementary material has been organized into the following sections:
\begin{itemize}
\item Section~\ref{contribution}: Technical Contribution and Novelty
\item Section~\ref{SecA}: Details on S.O.T.A NeRF Editing Methods
\item Section~\ref{algo}: LatentEditor Algorithm
\item Section~\ref{SecD}: Adapter Design Details
    \item Section~\ref{SecF}: Implementation Details
    \item Section~\ref{secG}: Additional Results
     \item Section~\ref{secH}: User Study
     \item Section~\ref{sec:broader_impact}: Broader Impact
     
\end{itemize}

\section{Technical Contribution and Novelty}
\label{contribution}
In the proposed LatentEditor framework, our primary focus is on enabling local editing of Neural Radiance Fields (NeRF). To facilitate this, we introduce a novel delta module that assigns delta scores within the latent space. This approach forms the basis of our distinctive technique for training NeRF on real-world 3D scenes within the latent space, capitalizing on the benefits of latent masks for precise editing.\\
While the concept of training Neural Radiance Fields (NeRF) in the latent space of Latent Diffusion Models was initially introduced by Latent-NeRF~\cite{metzer2023latent}, their focus was limited to generating 'virtual' 3D objects from textual conditions, rather than real-world 3D data. Our approach, in contrast, represents a novel advancement by training latent space NeRF for authentic 3D scenes. We achieve this by embedding RGB images of real scenes as latent representations and introducing a unique refinement module and camera parameters pre-conditioning to enhance reconstruction performance. Training NeRF in the latent space provides benefits like diminished training load and improved editability, leveraging the direct use of output rendered in Latent Diffusion Model (LDM) frameworks.

Our work marks a substantial contribution to the field, demonstrating a significant technical leap in the realm of NeRF research, which has predominantly been focused on training with natural RGB images. This expansion into real 3D scene reconstruction using latent space methodologies is a notable stride in advancing NeRF capabilities. 

 There exist several NeRF backbone models that facilitate faster training. However, editing a NeRF model in tandem with a Diffusion model necessitates rendering high-resolution imagery (e.g., 512x512) and computing gradients relative to it, which incurs considerable GPU memory use. Our model, trained on lower-dimension data (e.g., 64x64), circumvents this challenge by significantly reducing memory requirements for NeRF fine-tuning. By addressing the core issue of training efficiency through the reduction in data dimensions, our approach realizes a more resource-efficient training process compared to conventional NeRF models operating in image space.
 
In summary, LatentEditor exhibits two key attributes for effective NeRF editing: $(i)$ enhanced speed and resource efficiency in the NeRF editing process, and $(ii)$ the ability to perform local editing while maintaining the integrity of the background.

\section{Related Work on NeRF Local Editing}\label{SecA}

Within the domain of neural radiance fields, InstructNerf2Nerf (IN2N)~\cite{haque2023instruct} represents a significant step forward by utilizing 2D image translation models, notably InstructPix2Pix (IP2P)~\cite{brooks2023instructpix2pix}, for adjusting 2D image features to optimize NeRF training, under the guidance of textual prompts. However, the heavy reliance of IN2N on IP2P for processing NeRF training data occasionally results in over-editing of scenes. To mitigate this, the DreamEditor framework~\cite{zhuang2023dreameditor} was proposed, introducing a mesh-based neural field method. This strategy enables the conversion of 2D masks to 3D editing zones, allowing for targeted local modifications while preventing unnecessary geometric changes when only altering the visual appearance.

\paragraph{DreamEditor Analysis.} DreamEditor~\cite{zhuang2023dreameditor} employs pre-computed editing masks in tandem with text prompts to streamline the editing workflow. Although DreamEditor claims the use of a diffusion model~\cite{rombach2022high} for mask generation, the specifics of this process remain ambiguous. A thorough examination of their implementation~\footnote{\url{https://github.com/zjy526223908/dreameditor}} reveals that DreamEditor actually utilizes a pre-trained segmentation model for producing local editing masks, which are included in their provided dataset~\footnote{\url{https://drive.google.com/drive/folders/12_PlC9cd8ZPg3tXrxJ1m-HKAGYDiahjc}}. Moreover, DreamEditor faces challenges like prolonged training duration and limitations to specific scene types. Moreover, the segmentation models~\cite{luddecke2022image,kirillov2023segment} often fail to accurately interpret the editing prompts, necessitating object specification in a singular format for segmentation.

Contrastingly, our method innovates by integrating latent space training in NeRF with a novel delta module that utilizes diffusion models for mask generation. This approach enables accurate local editing without the need for additional guidance. The effectiveness of our method is demonstrated through comparative analyses presented in the main paper and the provided videos along with the \textit{supplementary material} highlighting our framework's enhanced capabilities in local editing compared to other text-driven NeRF methodologies.
\section{LatentEditor Algorithm}
\label{algo}
In this section, we introduce the algorithm underlying our proposed LatenEditor framework.
\begin{algorithm}
\caption{LatentEditor for NeRF Editing}

\begin{algorithmic}[1]
\State \textbf{Input:} Multi-view image dataset $\mathbf{I}$, text prompt $C_e$, editing rate $\upsilon$, training iteration $i$
\State \textbf{Output:} Edited NeRF model $F_{\theta}$ and Adapter $F_{\Theta}$

\State // Model Initialization
\State Encode $\mathbf{I}$ using encoder $\mathcal{E}$ to obtain $\mathbf{Z} := \{{z^n} = \mathcal{E}({{I^n}})\}_{n=1}^N$
\State Train  $F_{\theta}$ and $F_{\Theta}$ using loss $\mathcal{L}_{T}$ with $\mathbf{Z}$ 

\State // NeRF Editing
\For{$i = 1$ \textbf{to} $\# of iterations$}
    \If{$i / \upsilon == 0$}
        \State Generate noisy latent ${z}^n_{\Delta t}$ 
        \State Calculate delta map $\Delta_\varepsilon$
        \State Generate binary mask $M$ from $\Delta_\varepsilon$
    
        \State Generate another noisy version $\tilde{z}^n_{t}$
        \State Compute edited latent ${z}^n_e$ 
        \State Update latent $z^n$ with edited latent $z_e^n$
     \EndIf
\EndFor

\State // Optimizing NeRF Editing
\State Update $F_{\theta}$, $F_{\Theta}$  with modified set~$\mathbf{Z_e}$ using loss $\mathcal{L}_{T}$

\end{algorithmic}
\end{algorithm}

\section{Adapter Design Details}
\label{SecD}
As previously discussed, the original idea of translating NeRF into latent space was pioneered by Latent-NeRF~\cite{metzer2023latent}, with an emphasis on creating 'virtual' 3D assets. The authors of~\cite{metzer2023latent} further enhanced this model by performing fine-tuning in the pixel space, enabling the NeRF model to function directly with RGB values. This was achieved by transforming a NeRF trained in latent space into one operating in RGB space through a linear layer. Such refinement becomes essential due to the unique architecture of the encoder and decoder in Stable Diffusion~\cite{rombach2022high}, which utilize ResNet blocks and self-attention layers, leading to intricate interactions among pixel values during the conversion from image to latent space. 

As we aim to train NeRF in latent space, converting images to latents creates a misalignment between latent and image pixel values, as NeRF's rendering method independently computes pixel values through an MLP, devoid of inter-pixel interaction.

In response to this challenge and to better adapt our latent editing approach to 3D real-world scenes, we have developed a specific refinement module. This module is distinguished by a trainable adapter that eliminates the necessity of transitioning from latent space to RGB for refinement. While adapters have been widely studied in the context of vision transformers~\cite{chen2022vision,tomanek2021residual, khalid2023cefhri}, our adapter is uniquely tailored to incorporate the inter-pixel interactions inherent to ResNet and self-attention mechanisms. This integration substantially improves the reconstruction quality. The designed adapter has nearly 0.28 million parameters.

\section{Implementation Details}
\label{SecF}
We set $\lambda_{r} = 0.80$, $\lambda_{f}= 0.1$, and $\lambda_{p}= 0.1$  for the initial 2500 steps in model initialization. For the remaining steps,  we set $\lambda_{r} = 0.75$,$\lambda_{p}= 0.25$ and $\lambda_{f}= 0.0$. During Editing, we don't refine the camera parameters, so $\lambda_{p}$ is always 0. We set $\lambda_{r} = 0.90$, and $\lambda_{p}= 0.1$ for the initial 400 steps, and thereafter, set $\lambda_{p} = 0$, and $\lambda_{r} = 1.0$

For NeRF editing, we use $s_T = 7.5$ and $s_I=1.5$ in all of our experiments. For the threshold, we set $\mu = 0.45$. 
For all other experimental settings and model configurations, we use the default settings of ~\cite{haque2023instruct}.
\subsection{Model Initialization - Efficiency Analysis}
\label{eff1}
The primary advantage of training the Nerfacto model~\cite{tancik2023nerfstudio} in latent space, rather than pixel space, lies in the significantly lower resolution of latent representations compared to RGB images. Specifically, the Stable Diffusion~\cite{rombach2022high} VAE used in our study downscales an input image of resolution W$\times$H$\times$3 to latents of size H/8$\times$ W/8 $\times$ 4. As a result, the latent Nerfacto model processes 64 times fewer rays than its RGB counterpart. Theoretically, this reduction allows for training with a batch size ($\#$ of rays) 64 times smaller while keeping the number of model initialization training iterations constant (30,000 iterations, as in IN2N~\cite{haque2023instruct}), leading to a substantially smaller memory footprint. Alternatively, one could maintain the same batch size as the RGB model (in IN2N~\cite{haque2023instruct}) and reduce the number of iterations by a factor of 64 to achieve effective latent output. However, in our experiments, we chose a batch size of 1,024 and trained for 30,000 iterations. Under this setup, our models require 1,236MB of GPU memory (including memory for the Refinement Adapter), in contrast to the 2,086MB needed by the RGB model as indicated in Table~\ref{tab:gpu_memory_usage}. In terms of time efficiency, our complete training pipeline, which includes latent generation, training the latent Nerfacto, and training the Refinement Adapter (1,000 iterations), is completed in 758 seconds. These results were obtained using the “fangzhou-small” scene from IN2N~\cite{haque2023instruct} dataset on NVIDIA RTX 4090 GPU. 
However, in our analysis, we observe that LatentEditor can achieve similar results with the initial training of 15,000 iterations bringing the total training time below 400 seconds, significantly faster than the 745 seconds required by the Nerfacto model in IN2N~\cite{haque2023instruct}.

\begin{table}[htb]
\centering
\caption{\footnotesize Comparison of GPU Memory Consumption between IN2N and the proposed LatentEditor Using the Nerfacto Model during model initialization stage}
\resizebox{0.85\textwidth}{!}{
\begin{tabular}{lccc}
\hline
\textbf{Method}          & \textbf{Nerfacto Training} & \textbf{Adapter Training} & \textbf{Total Memory} \\ \hline
\textbf{IN2N~\cite{haque2023instruct}}  & 2086 MB          & -                & 2086 MB        \\
\textbf{LatentEditor} & 1158 MB           & 78 MB            & 1236 MB     \\ \hline
\end{tabular}
}

\label{tab:gpu_memory_usage}

\end{table}

\begin{figure*}[htbp!]
\centering
    \includegraphics[width=1\linewidth, trim={0cm 4.5cm 0cm 3cm},clip]{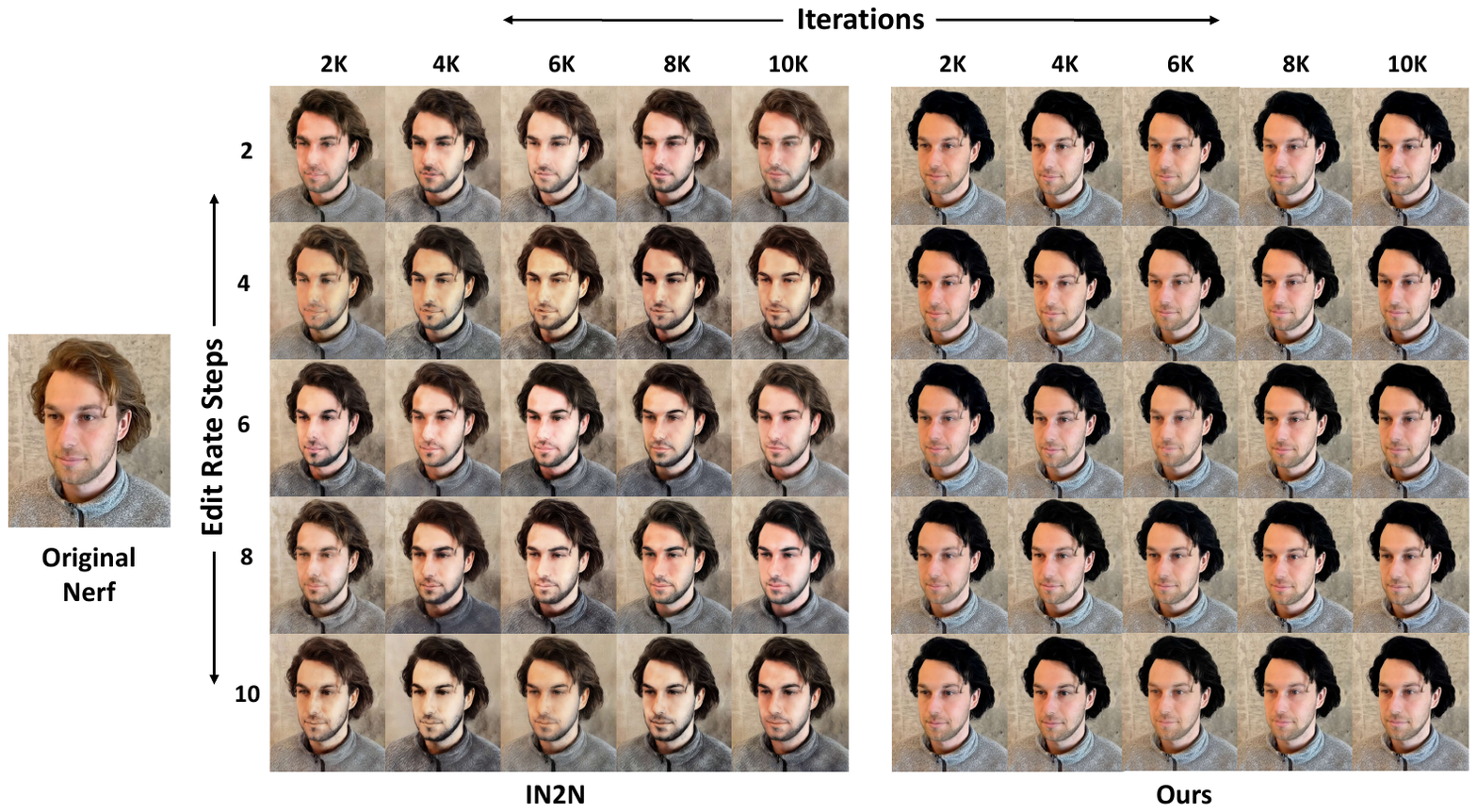}
    
    \caption{\footnotesize This figure provides a comparative analysis between \textit{LatentEditor} and IN2N~\cite{haque2023instruct}, focusing on computational efficiency across different editing rates. The aim is to reduce the number of required training iterations for higher editing rates (less frequent IP2P calls). Extending the comparison presented in ablations of the main paper, where we used a default editing rate of 10, this figure explores editing rates ranging from 2 to 10 and documents the outcomes for various numbers of fine-tuning iterations. \textit{LatentEditor} demonstrates effective editing within approximately 2000 iterations, while IN2N struggles to achieve comparable results even after 10,000 iterations.}

\label{fig:ablations_long}
\end{figure*}

\subsection{NeRF Editing - Efficiency Analysis}
When considering the "fangzhou-small" scene from the IN2N dataset~\cite{haque2023instruct}, the batch size set during editing is 16,384, corresponding to the original RGB image dimensions of 512$\times$320 in the IN2N implementation. In this configuration, the NeRF model alone demands approximately 6,778 MB of GPU memory, excluding the additional memory required by the diffusion model. On the other hand, LatentEditor, leveraging latent dimensions of 64$\times$40 for the "fangzhou-small" image latents , permits a reduced batch size of 2,560. This adjustment significantly reduces the NeRF model's memory usage to 1,766 MB.

In addition, LatentEditor requires less number of fine-tuning iterations.
Figure~\ref{fig:ablations_long} presents a comparative analysis of \textit{LatentEditor} and IN2N~\cite{haque2023instruct} on the "face" scene of the IN2N dataset~\cite{haque2023instruct}, examining computational efficiency across various editing rates. This figure extends the analysis from the ablation study in the main paper, which used a default editing rate of 10, by exploring rates from 2 to 10 and recording the number of fine-tuning iterations. \textit{LatentEditor} efficiently achieves editing goals within about 2000 iterations, in contrast to IN2N, which requires over 10,000 iterations for similar results. This ablation study highlights the superior efficiency of \textit{LatentEditor} in terms of required training iterations, achieving a 5 to 10-fold reduction compared to IN2N, despite the increased cost per step due to multiple denoising stages. Consequently, \textit{LatentEditor} is \textbf{5-8} times faster in NeRF editing than IN2N.

\subsection{Efficient Rendering}
Using the LLFF dataset as a benchmark, Nerfacto requires an average of 2.85 seconds to render a single image at 512 resolution. In comparison, our model, which first renders a latent feature map and then decodes it to an RGB image, takes only an average of 1.35 seconds per image. This efficiency gain, amounting to a 2.1-fold reduction in inference time, is attributed to the training in latent space and the lowered resolution employed by LatentEditor.
\section{Additional Results}
\label{secG}

\begin{figure*}[htbp!]
\centering
    \includegraphics[width=1\linewidth, trim={2.4cm 4.6cm 4cm 5cm},clip]{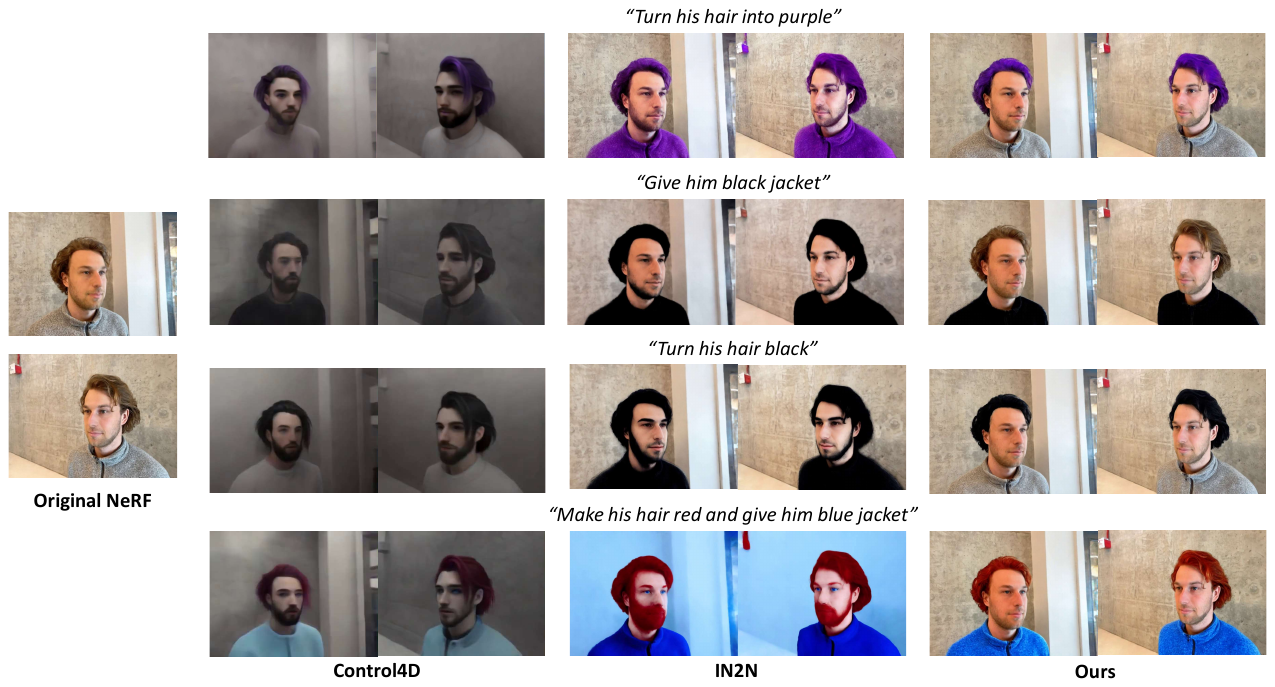}
    \caption{\footnotesize This figure presents a comparison of LatentEditor, IN2N, and Control4D in terms of local and multi-attribute editing. The comparison reveals that Control4D modifies the original scene content. IN2N, while editing, inadvertently alters unintended regions, such as changing the beard and jacket color when only a hair color change is intended. In contrast, LatentEditor successfully accomplishes the desired editing outcomes as specified by the given prompts.
}
\label{fig:control4D}
\end{figure*}
\begin{figure}[t!]

\centering
    \includegraphics[trim={7cm 5.7cm 10cm 3.3cm}, clip, width=\linewidth]{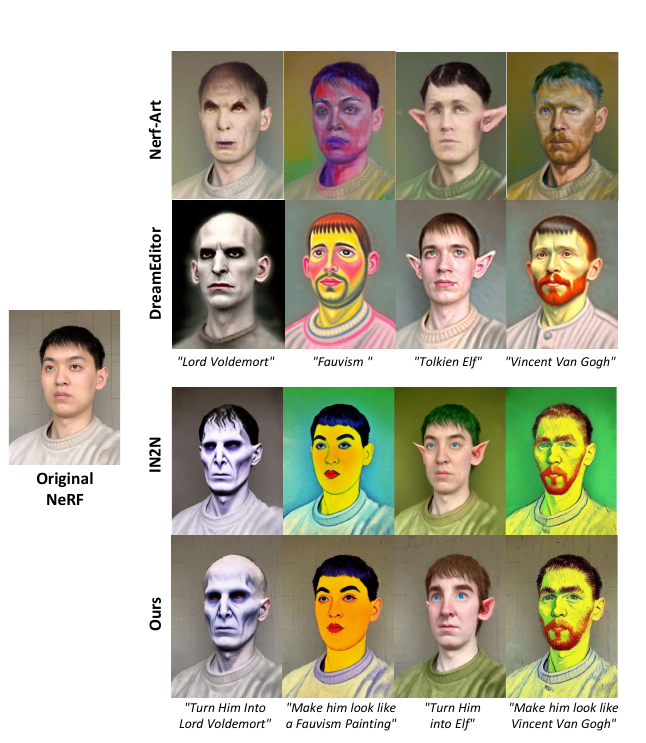}

\caption{\footnotesize This figure compares LatentEditor with Nerfart, DreamEditor, and IN2N. LatentEditor stands out by achieving the targeted style transfer outcomes with a significantly lower number of training iterations (approximately 3000), showcasing a notable efficiency advantage over IN2N (requires about 18000 iterations).}

\label{fig:comparison_style}
\end{figure}


We further present the qualitative comparisons with Control4D~\cite{shao2024control4d} and DreamEditor~\cite{zhuang2023dreameditor} in Figure~\ref{fig:control4D}. Figure~\ref{fig:control4D} illustrates the contrasts in local and multi-attribute editing among LatentEditor, IN2N, and Control4D. As the Control4D~\cite{shao2024control4d} code is not publicly accessible, we developed an in-house implementation to assess the impact of GAN guidance.
 Control4D tends to alter the original scene content, while IN2N inadvertently changes unintended areas, like beard and jacket color when only a hair color change is intended. Conversely, LatentEditor precisely achieves the editing objectives as per the prompts, demonstrating its superior accuracy in targeted modifications.

 \begin{table}[t]
\centering
\caption{\footnotesize \textbf{Multi-attribute Editing.} Quantitative evaluation of scene edits in terms of text alignment and frame consistency in CLIP space where our approach demonstrates the highest consistency. }
\resizebox{0.95\textwidth}{!}{
\begin{tabular}{lccc}
\hline
\textbf{Method} &
  \textbf{\begin{tabular}[c]{@{}c@{}}CLIP Text-Image\\
Direction Similarity$\uparrow$\end{tabular}} &
  \textbf{\begin{tabular}[c]{@{}c@{}}CLIP Direction\\
 Consistency$\uparrow$\end{tabular}} &
  \textbf{\begin{tabular}[c]{@{}c@{}}Edit\\ PSNR$\uparrow$\end{tabular}} \\ \hline
\textbf{NeRF-Art}~\cite{wang2023nerf} & 0.2007 & 0.8564 & 16.54 \\
\textbf{Control4D~\cite{shao2024control4d}} & 0.1945 & 0.8664 & 14.45 \\
\textbf{DreamEditor~\cite{zhuang2023dreameditor}} & 0.2134 & 0.8702 & 18.33 \\
\textbf{IN2N}~\cite{haque2023instruct} & 0.2257 & 0.8846 & 20.07 \\
\textbf{Collaborative Score Distillation}~\cite{kim2023collaborative} & 0.2134 & 0.8755 & 19.46 \\
\textbf{ViCA-NeRF}~\cite{dong2024vica} & 0.2112 & 0.8636 & 18.76 \\
\textbf{Ours}             & \textbf{0.2611} & \textbf{0.9347} & \textbf{24.45} \\ \hline
\end{tabular}
}

\label{tab:multi}
\end{table}

Moreover, Figure~\ref{fig:comparison_style} compares LatentEditor with NeRF-Art~\cite{wang2023nerf}, DreamEditor~\cite{zhuang2023dreameditor}, and IN2N~\cite{haque2023instruct}, focusing on style transfer capabilities. LatentEditor distinguishes itself by achieving the desired style transfer results in significantly fewer training iterations, about 3000, compared to IN2N, which requires approximately 18000 iterations. This efficiency in obtaining high-quality outcomes with fewer iterations highlights the practical utility of LatentEditor in style transfer applications.

\subsection{Quantitative Evaluation Metrics}
In our quantitative evaluation, we utilize three metrics: the \textit{CLIP Text-Image
Direction Similarity}~\cite{gal2021stylegan}, the \textit{CLIP Direction Consistency Score}~\cite{haque2023instruct}, and~\textit{ Edit PSNR}. 
The CLIP Directional Score evaluates the alignment between the textual caption changes and the corresponding image transformations. In contrast, the CLIP Consistency Score assesses the cosine similarity between the CLIP embeddings of consecutive frames along a novel camera trajectory. The Edit PSNR metric evaluates the cosine similarity and PSNR (Peak Signal-to-Noise Ratio) for each rendered view, comparing the edited NeRF against the input NeRF. This comparison demonstrates that our method maintains greater consistency in the edited scene with respect to the original input scene.

 \begin{table}[tb]
\centering
\caption{\footnotesize \textbf{Single-attribute Editing.} Quantitative evaluation of scene edits with View-Consistent Editing Frameworks}
\resizebox{0.95\textwidth}{!}{
\begin{tabular}{lccc}
\hline
\textbf{Method} &
  \textbf{\begin{tabular}[c]{@{}c@{}}CLIP Text-Image\\
Direction Similarity$\uparrow$\end{tabular}} &
  \textbf{\begin{tabular}[c]{@{}c@{}}CLIP Direction\\
 Consistency$\uparrow$\end{tabular}} &
  \textbf{\begin{tabular}[c]{@{}c@{}}Edit\\ PSNR$\uparrow$\end{tabular}} \\ \hline
\textbf{Collaborative Score Distillation}~\cite{kim2023collaborative} & 0.2417 & 0.9236 & 22.17 \\
\textbf{ViCA-NeRF}~\cite{dong2024vica} & 0.2452 & 0.8966 & 21.23 \\
\textbf{Ours}             & \textbf{0.2661} & \textbf{0.9387} & \textbf{25.15} \\ \hline
\end{tabular}
}

\label{tab:consistent}
\end{table}
In the main paper, we report the single-attribute editing quantitative results as IN2N~\cite{haque2023instruct} doesn't support multi-attribute editing inherently. However, we report the multi-attribute editing perfromance of various NeRF-Editing methods in Table~\ref{tab:multi}. In addition, we also compare our method with view-consistent 3D editing methods accepted in NeurIPS, 2024 in Table~\ref{tab:consistent} using their open source code.

\subsection{Comparing with Mask+IP2P}
We further evaluate LatentEditor by combining a pre-trained segmentation model SAM~\cite{kirillov2023segment} with IN2N~\cite{haque2023instruct} and IP2P~\cite{brooks2023instructpix2pix}. The quantitative results are presented in Table~\ref{tab:mask}.

\begin{table}[htb]
\centering
\caption{\footnotesize Quantitative evaluation by using pre-trained segmentation model with IN2N and IP2P.  }
\resizebox{0.85\textwidth}{!}{
\begin{tabular}{lccc}
\hline
\textbf{Method} &
  \textbf{\begin{tabular}[c]{@{}c@{}}CLIP Text-Image\\
Direction Similarity$\uparrow$\end{tabular}} &
  \textbf{\begin{tabular}[c]{@{}c@{}}CLIP Direction\\
 Consistency$\uparrow$\end{tabular}} &
  \textbf{\begin{tabular}[c]{@{}c@{}}Edit\\ PSNR$\uparrow$\end{tabular}} \\ \hline
\textbf{IP2P~\cite{brooks2023instructpix2pix}+SAM~\cite{kirillov2023segment} } & 0.2285 & 0.9172 & 20.15 \\
\textbf{IN2N~\cite{haque2023instruct}+SAM~\cite{kirillov2023segment}} & \textbf{0.2865} & 0.9251 &24.89 \\
\textbf{LatentEditor}             & 0.2661 & \textbf{0.9387} & \textbf{25.15} \\ \hline
\end{tabular}
}

\label{tab:mask}
\end{table}

\subsection{Additional Ablations}
In the ablations section of our study, we meticulously dissect the impact of distinct components within our framework to ascertain their individual contributions towards the overall performance. Specifically, we conduct a detailed examination by selectively omitting key elements from our model and observing the resultant effects on its efficacy. Among these components, significant attention is directed towards the "Delta Module" and "Camera Parameters Alignment," both of which are hypothesized to play pivotal roles in our framework's functionality. To present our findings in a structured and comprehensible manner, we compile the results derived from these investigations into Table~\ref{tab:ab_ex}. This table juxtaposes the performance metrics of our complete model against versions devoid of the Delta Module and Camera Parameters Alignment, respectively, providing clear insights into the value added by each component. Through this analytical approach, we aim to highlight the essentiality of these mechanisms in achieving the high-quality outcomes our framework is designed to deliver.

\begin{table}[htb]
\centering
\caption{\footnotesize \textbf{Additional Ablations.} The results are presented to evaluate the performance of the proposed framework without optimization of camera parameters in latent space and without applying the the delta module. }
\resizebox{0.98\textwidth}{!}{
\begin{tabular}{lccc}
\hline
\textbf{Adapter Type} &
  \textbf{\begin{tabular}[c]{@{}c@{}}CLIP Text-Image\\
Direction Similarity$\uparrow$\end{tabular}} &
  \textbf{\begin{tabular}[c]{@{}c@{}}CLIP Direction\\
 Consistency$\uparrow$\end{tabular}} &
  \textbf{\begin{tabular}[c]{@{}c@{}}Edit\\ PSNR$\uparrow$\end{tabular}} \\ \hline
\textbf{w/o Delta Module } & 0.2585 & 0.9272 & 23.85 \\
\textbf{w/o Camera Parameters Alignment} & 0.2003 & 0.8981 & 21.89 \\
\textbf{Ours}             & \textbf{0.2661} & \textbf{0.9387} & \textbf{25.15} \\ \hline
\end{tabular}
}

\label{tab:ab_ex}
\vspace{-4mm}
\end{table}
\section{User Study Details}
\label{secH}
To assess the perceptual preferences of human subjects towards edited NeRF renderings, we conducted a comprehensive user study. This study involved rendering images from 8 distinct scenes taken from the LLFF, IN2N, and NeRFstudio datasets. Feedback was solicited from 500 participants, whose ages ranged from 18 to 68 years. During the study, each participant was shown a series of multi-view renderings generated by our model and various baseline models. The presentation order of these renderings was randomized to ensure unbiased feedback.

Participants were asked to rate each rendering using a preference scoring survey, with the scoring scale ranging from 1 (very low) to 10 (very high), encompassing five distinct rating options. The study aimed to evaluate three key aspects of the edited NeRF renderings: 

\begin{enumerate}
    \item \textbf{Text Fidelity}: Whether the image accurately reflects the target text condition.
    \item \textbf{Content Preservation}: The model's effectiveness in precisely editing the targeted object.
    \item \textbf{Scene Consistency}: The degree to which the 3D scenes maintain consistency across different views.
\end{enumerate}
 
The aggregated results of this user study are presented in the main paper. Our method demonstrated superior performance in terms of text fidelity and content preservation, achieving the highest scores in these categories. Additionally, it ranked second in scene consistency, underlining its effectiveness in maintaining view coherence. Overall, our approach surpassed the baseline models in terms of perceptual quality, as reflected in the user feedback. 

\vspace{-2mm}
\section{Broader Impact}
\label{sec:broader_impact}
\vspace{-2mm}
The development and deployment of LatentEditor carry substantial implications for both academic research and practical applications, highlighting a pivotal advancement in computer graphics and neural rendering. This technology simplifies the process of 3D scene editing, democratizing access to high-quality content creation for artists, designers, and content creators. By enabling precise, text-driven modifications without requiring extensive technical knowledge, LatentEditor fosters creativity and innovation across digital media, entertainment, and virtual reality domains.

The societal benefits of LatentEditor extend to enhancing educational software and training simulations, where the ability to modify 3D scenes easily can lead to more realistic and context-specific scenarios. This is particularly valuable in fields such as medicine, architecture, and emergency response training. Furthermore, the technology's impact on gaming and virtual reality promises richer, more immersive user experiences, with its efficient and localized editing capabilities potentially making high-quality content more accessible to smaller development teams. However, these advancements come with ethical considerations. The ease and precision of LatentEditor's editing capabilities raise concerns about the potential for creating misleading or deceptive content, such as deepfakes. It necessitates the development of ethical guidelines and the use of detection technologies to mitigate such risks. Additionally, ensuring privacy and securing consent for the use of personal likenesses or private locations in edited content is paramount, requiring established guidelines and legal frameworks. 

Looking forward, LatentEditor opens new avenues for research in neural rendering and the efficiency and realism of 3D content creation. Extending the framework to support dynamic scenes and real-time editing could further bridge the gap between virtual and real-world interactions. Integrating LatentEditor with other AI-driven tools could unlock novel applications in film production, architectural visualization, and augmented reality, offering users novel ways to interact with and modify their environments. In the realm of augmented reality (AR), LatentEditor's potential is equally transformative. By facilitating the creation and modification of 3D content with ease and precision, it can greatly enhance the development of AR applications, ranging from interactive learning tools to immersive retail experiences. This could lead to a significant shift in how users interact with digital information in their physical environment, making AR/VR devices more versatile and useful across a broad spectrum of activities.
{
    \small
    \bibliographystyle{splncs04}
    \bibliography{egbib}
}